\documentclass{article}

    \usepackage[final]{neurips_2022}

\usepackage[utf8]{inputenc} 
\usepackage[T1]{fontenc}    
\usepackage{hyperref}       
\usepackage{url}            
\usepackage{booktabs}       
\usepackage{amsfonts}       
\usepackage{nicefrac}       
\usepackage{microtype}      
\usepackage{xcolor}         

\usepackage{enumitem}
\usepackage{booktabs}  
\usepackage{graphicx}
\usepackage{caption}
\usepackage{subcaption}
\usepackage{wrapfig}
\usepackage{amssymb}
\usepackage{amsmath}
\DeclareMathOperator*{\argmax}{arg\,max}
\DeclareMathOperator*{\argmin}{arg\,min}
\usepackage{makecell}

\usepackage{multirow}
\usepackage{placeins}
\usepackage[normalem]{ulem}

\usepackage{bbm}
\usepackage{amsthm}

\newtheorem{definition}{Definition}[section]
\newtheorem{example}{Example}[section]
\newtheorem{theorem}{Theorem}[section]
\newtheorem{lemma}[theorem]{Lemma}

\newtheorem{remark}{Remark}[section]
\newtheorem{assumption}{Assumption}[section]

\newcommand{\smallsection}[1]{\textbf{#1.~~~~}}

\newcommand{\noise}[1]{implicit label noise\xspace}
\newcommand{\Noise}[1]{Implicit label noise\xspace}

\title{Label Noise in Adversarial Training: A Novel Perspective to Study Robust Overfitting}

\author{%
Chengyu Dong \\
University of California, San Diego\\
\texttt{cdong@eng.ucsd.edu}\\
\And
Liyuan Liu\\
Microsoft Research\\
\texttt{lucliu@microsoft.com}\\
\And
Jingbo Shang\\
University of California, San Diego\\
\texttt{jshang@eng.ucsd.edu}\\
}

\begin{document}

\maketitle

\begin{abstract}


We show that label noise exists in adversarial training. 
Such label noise is due to the mismatch between the true label distribution of adversarial examples and the label inherited from clean examples 
 -- the true label distribution is distorted by the adversarial perturbation, but is neglected by the common practice that inherits labels from clean examples. 
Recognizing label noise sheds insights on the prevalence of robust overfitting in adversarial training, and explains its intriguing dependence on perturbation radius and data quality. 
Also, our label noise perspective aligns well with our observations of the epoch-wise double descent in adversarial training. 
Guided by our analyses, we proposed a method to automatically calibrate the label to address the label noise and robust overfitting. 
Our method achieves consistent performance improvements across various models and datasets without introducing  new hyper-parameters or additional tuning.

\end{abstract}

\section{Introduction}



Adversarial training~\citep{Goodfellow2015ExplainingAH, Huang2015LearningWA, Kurakin2017AdversarialML, Madry2018TowardsDL} is known as one of the most effective ways~\citep{Athalye2018ObfuscatedGG, Uesato2018AdversarialRA} to enhance the adversarial robustness of deep neural networks~\citep{Szegedy2014IntriguingPO, Goodfellow2015ExplainingAH}. 
It augments training data with adversarial perturbations to prepare the model for adversarial attacks. 
Despite various efforts to generate more effective adversarial training examples~\citep{Ding2020MaxMarginA, Zhang2020AttacksWD},
the labels assigned to them attracts little attention. 
As the common practice, the assigned labels of adversarial training examples are simply inherited from their clean counterparts. 

\begin{figure}[tp]
  \centering
  \includegraphics[width=1.00\textwidth]{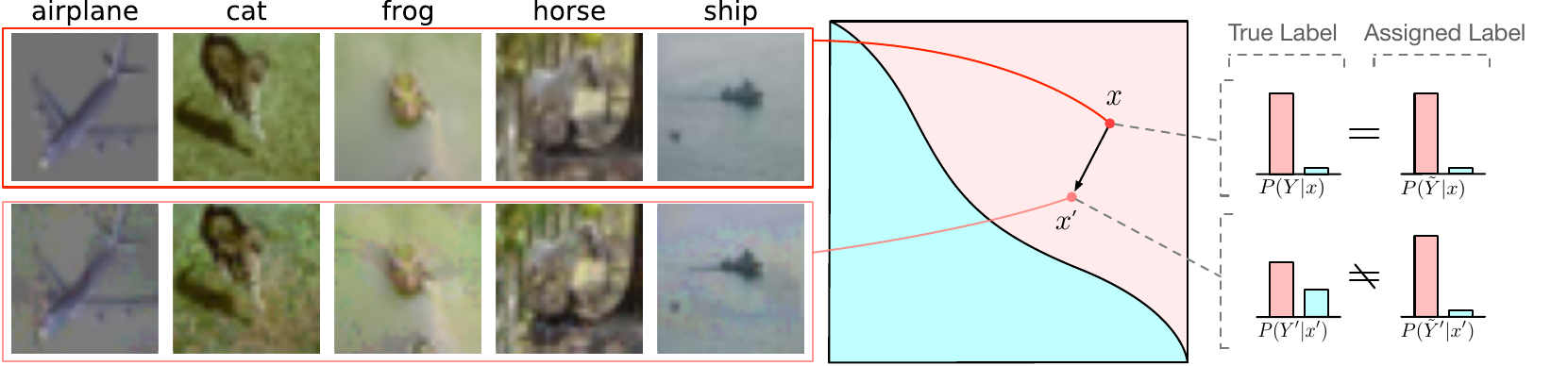}
  \vspace{-1ex}
  \caption{Illustration of the origin of label noise in adversarial training. 
  The adversarial perturbation causes a mismatch between the true label distributions of clean inputs $x$ and their adversarial examples $x'$. Such a distribution mismatch is however neglected by the labels assigned to adversarial examples in the common practice of adversarial training, resulting in label noise implicitly.
  }
 \vspace{-2ex}
\label{fig:illustration}
\end{figure}

In this paper, we argue that the existing labeling practice of the adversarial training examples introduces label noise implicitly, since adversarial perturbation can distort the data semantics~\citep{Tsipras2019RobustnessMB, Ilyas2019AdversarialEA}.
For example, as illustrated in Figure~\ref{fig:illustration}, even with a slight distortion of the data semantics (e.g., more ambiguous), the label distribution of the adversarially perturbed data may not match the label distribution of the clean counterparts.
Such distribution shift is neglected when assigning labels to adversarial examples, which are directly copied from the clean counterparts.
We observe that distribution mismatch caused by adversarial perturbation along with improper labeling practice will cause \emph{label noise
in adversarial training}.

It is a mysterious and prominent phenomenon that the robust test error would start to increase after conducting adversarial training for a certain number of epochs~\citep{Rice2020OverfittingIA}, and our label noise perspective provides an adequate explanation for this phenomenon. 
Specifically, from a classic bias-variance view of model generalization, label noise that implicitly exists in adversarial training can increase the model variance~\citep{Yang2020RethinkingBT} and thus make the overfitting much more evident compared to standard training.
Further analyses of label noise in adversarial training also explain the intriguing dependence of robust overfitting on the perturbation radius~\citep{Dong2021ExploringMI} and data quality~\citep{Dong2021DataPF} presented in the literature.

\begin{wrapfigure}{r}{6cm} 
  \centering
  \includegraphics[width=1.0\linewidth]{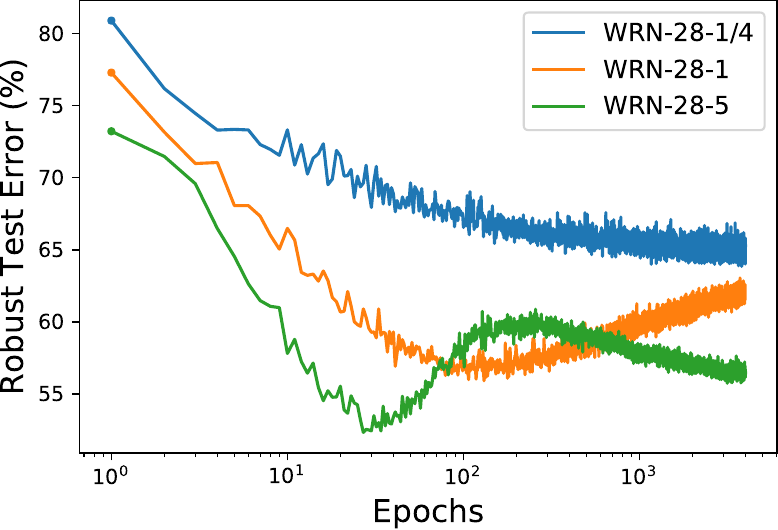}
  \caption{Robust overfitting can be viewed as an early part of the epoch-wise double descent. 
  We employ PGD training~\citep{Madry2018TowardsDL} on CIFAR-10~\citep{Krizhevsky2009LearningML} with Wide ResNet (WRN)~\citep{Zagoruyko2016WideRN} and a fixed learning rate. 
  WRN-$28$-$k$ refers to WRN with depth $28$ and widen factor $k$.
  }
\label{fig:intro}
\vspace{-3mm}
\end{wrapfigure}

Providing the label noise in adversarial training, one can further expect the existence of double descent based on the modern generalization theory of deep neural networks.
\emph{Epoch-wise double descent} refers to the phenomenon that the test error will first decrease and then increase as predicted by the classic bias-variance trade-off, but it will decrease again as the training continues. Such phenomenon is only reported in standard training of deep neural networks, often requiring significant label noise in the training set~\citep{Nakkiran2020DeepDD}.
As the label noise intrinsically exists in adversarial training, such epoch-wise double descent phenomenon also emerges when the training goes longer.
Indeed, as shown in Figure~\ref{fig:intro}, for a relatively large model such as WRN-28-5, on top of the existing robust overfitting phenomenon, the robust test error will eventually decrease again \emph{after $1,000$ epochs}. Following \cite{Nakkiran2020DeepDD}, we further experiment different model sizes. One can find that a medium-sized model will follow a classic U-curve, which means only overfitting is observed; and the robust test error for a small model will monotonically decrease. These are well aligned with the observations in standard training regime.
This again consolidates our understanding of label noise in adversarial training.

In light of our analyses, we design a theoretically-grounded method to mitigate the label noise in adversarial training automatically.
The key idea is to resort to an alternative labeling of the adversarial examples. 
We show that the predictive label distribution of an adversarially trained probabilistic classifier can approximate the true label distribution with high probability. Thus it can be utilized as a better labeling of the adversarial examples and provably reduce the label noise. We also show that with proper temperature scaling and interpolation, such predictive label distribution can further reduce the label noise.
This echoes the recent empirical practice of incorporating knowledge distillation~\citep{Hinton2015DistillingTK} into adversarial training~\citep{chen2021robust}. 
While previous works heuristically select fixed scaling and interpolation parameters for knowledge distillation, we show that it is possible to fully unleash the potential of knowledge distillation by automatically determining the set of parameters that maximally reduces the label noise, with a strategy similar to confidence calibration~\citep{Guo2017OnCO}. 
Such strategy can further mitigate robust overfitting to a minimal amount without additional human tuning effort.
Extensive experiments on different datasets, training methods, neural architectures and robustness evaluation metrics verify the effectiveness of our method.

In summary, our findings and contributions are: 1) we show that the labeling of adversarial examples in adversarial training practice introduces label noise implicitly; 2) we show that robust overfitting can be adequately explained by such label noise, and it is the early part of an epoch-wise double descent; 3) 2e show an alternative labeling of the adversarial examples can be established to provably reduce the label noise and mitigate the robust overfitting.

\section{Related Work}
\label{sect:related}

\smallsection{Robust overfitting and double descent in adversarial training}
Double descent refers to the phenomenon that overfitting by increasing model complexity will eventually improve test set performance~\citep{Neyshabur2017ExploringGI, Belkin2019ReconcilingMM}.
This appears to conflict with the robust overfitting phenomenon in adversarial training, where increasing model complexity by training longer will impair test set performance constantly after a certain point during training. It is thus believed in the literature that robust overfitting and epoch-wise double descent are separate phenomena~\citep{Rice2020OverfittingIA}. In this work we show this is not the complete picture by conducting adversarial training for exponentially more epochs than the typical practice.

A recent work also considers a different notion of double descent that is defined with respect to the perturbation size~\citep{Yu2021UnderstandingGI}. 
Such double descent might be more related to the robustness-accuracy trade-off problem~\citep{Papernot2016TowardsTS, Su2018IsRT, Tsipras2019RobustnessMB, Zhang2019TheoreticallyPT}, rather than the classic understanding of double descent based on model complexity.

\smallsection{Mitigate robust overfitting}
Robust overfitting hinders the practical deployment of adversarial training methods as the final performance is often sub-optimal. Various regularization methods including classic approaches such as $\ell_1$ and $\ell_2$ regularization and modern approaches such as cutout~\citep{Devries2017ImprovedRO} and mixup~\citep{Zhang2018mixupBE} have been attempted to tackle robust overfitting, whereas they are shown to perform no better than simply early stopping the training on a validation set~\citep{Rice2020OverfittingIA}. However, early stopping raises additional concern as the best checkpoint of the robust test accuracy and that of the standard accuracy often do not coincide~\citep{chen2021robust}, thus inevitably sacrificing the performance on either criterion. Various regularization methods specifically designed for adversarial training are thus proposed to outperform early stopping, including regularization the flatness of the weight loss landscape~\citep{Wu2020AdversarialWP, Stutz2021RelatingAR}, introducing low-curvature activation functions~\citep{Singla2021LowCA}, data-driven augmentations that adds high-quality additional data into the training~\citep{Rebuffi2021FixingDA} and adopting stochastic weight averaging~\citep{Izmailov2018AveragingWL} and knowledge distillation~\citep{Hinton2015DistillingTK}~\citep{chen2021robust}. These methods are likely to suppress the label noise in adversarial training, with the self-distillation framework (i.e. the teacher shares the same architecture as the student model) introduced by \citep{chen2021robust} as a particular example since introducing teacher's outputs as supervision is almost equivalent to the alternative labeling inspired by our understanding of the origin of label noise in adversarial training.


\section{Preliminaries}
\label{sect:notation}


    \smallsection{A statistic model of label noise~\citep{Frnay2014ClassificationIT}}
    Let $\mathcal{X} \subset \mathbbm{R}^d$ define the input space equipped with a norm $\|\cdot\|: \mathcal{X} \rightarrow \mathbbm{R}^+$ 
    and $\mathcal{Y} = [K] := \{1, 2,\ldots, K\}$ define the label space. 
    We introduce four random variables to describe noisy labeling process.
    Let $X \in \mathcal{X}$ denote the input, $Y \in \mathcal{Y}$ denote the true label of the input, $\tilde{Y} \in \mathcal{Y}$ denote the assigned label of an input provided by an annotator, and finally $E$ denote the occurrence of a label error by this annotator. $E$ is a binary random variable with value $1$ indicating that the assigned label is different from the true label for a given input, i.e., $E = \mathbf{1}(\tilde{Y} \ne Y)$. We study the case where the label error depends on both the input $X$ and the true label $Y$. 
    For a classification problem, a training set consists of a set of examples that are sampled as $\mathcal{D} =\{(x_i, \tilde{y}_i)\}_{i\in[N]}$. 
    
    \begin{definition}[Label noise]
    \label{definition:implicit-label-noise}
     We define label noise $p_e$ in a training set $\mathcal{D}$ as the empirical measure of the label error, namely $p_e(\mathcal{D}) = 1/N \sum_{i\in [N]} \mathbf{1}(\tilde{y}_i \ne y_i)$. 
    \end{definition}

    \begin{assumption}
        \label{assumption:clean-dataset}
        We assume the annotation of a clean dataset involves no label error, namely $P(E=1|Y=y, x) = 0$. This directly implies $P(\tilde{Y}|x) = P(Y|x)$ (see proof in the Appendix).
    \end{assumption}
    
    \begin{definition}[Data quality]
    \label{definition:data-quality}
    Given a training set $\mathcal{D}$, we define its data quality as $q(\mathcal{D}) = \mathbb{E}_{(x, y) \in \mathcal{D}} P(Y=y|x)$
    \end{definition}
    
    
    
    \smallsection{Adversarially augmented training set}
    Let $f: \mathcal{X} \rightarrow \mathcal{Y}$ be a probabilistic classifier and $f(\cdot)_j$ be its predictive probability at class $j$.
    The adversarial example of $x$ generated by $f$ is obtained by solving the maximization problem $x' = \argmax_{z \in \mathcal{B}_\varepsilon(x)}~\ell(f(z), y)$.
    Here $\ell$ can be a typical loss function such as cross-entropy. And $\mathcal{B}_\varepsilon (x)$ denotes the norm ball centered at $x$ with radius $\varepsilon$, i.e., $\mathcal{B}_\varepsilon (x) = \{z\in \mathcal{X}: \|z - x\| \le \varepsilon\}$.
    
    Following previous notations, we denote $Y'$ as a random variable representing the true label of $x'$ and $\tilde{Y'}$ as a random variable representing the assigned label of $x'$.
    We refer $\mathcal{D}' = \{(x', \tilde{y}')\}$ as the adversarially augmented training set.
    
    \smallsection{Adversarial training}
    Adversarial training can be viewed as a data augmentation technique that trains the parametric classifier $f_\theta$ on the adversarially augmented training set~\citep{Tsipras2019RobustnessMB}, namely
    \begin{equation}
        \label{eq:outer-minimization}
        \theta^* = \argmin_\theta \frac{1}{|\mathcal{D'}|}\sum_{(x', \tilde{y}') \in D'}~ \ell(f_\theta(x'), \tilde{y}').
    \end{equation}
\section{Label noise implicitly exists in adversarial training}
\label{sect:reason}
    
    In this section, we demonstrate the implicit existence of label noise in the adversarially augmented training set. We first consider a simple case where the adversarial perturbation is generated based on an ideal classifier that predicts the true label distribution. Under such a case we prove that the label noise in the adversarially augmented training set is lower-bounded. We then show that in realistic cases an adversarially trained classifier can approximate the true label distribution with high probability. Therefore, additional error terms will be required to lower bound the label noise.
    All proofs for the remainder of this paper are provided in the appendix.
    




    \subsection{When adversarial perturbation is generated by the true probabilistic classifier}
    \label{sect:reason-true}
    We first consider an ideal case where the adversarial perturbation is generated by the true probabilistic classifier $f(x):= P(Y|x)$, namely the classifier producing the true label distribution on any input $x$.
        
    

    \smallsection{The true label distribution is distorted by adversarial perturbation}
    We quantify the mismatch between two probability distributions using the \emph{total variation (TV) distance}.
    \begin{definition}[TV distance]
    \label{definition:total-variation-distance}
    Let $\mathcal{A}$ be a collection of the subsets of the label sample space $\mathcal{Y}$. The TV distance between two probability distributions $P(Y)$ and $P(Y')$ can be defined as $
       \|P(Y) - P(Y')\|_{\text{TV}} = \sup_{J\in \mathcal{A}} \left| P(Y\in J) - P(Y'\in J)\right| 
      $.
    \end{definition}
    
    We now show that adversarial perturbation generated by the true probabilistic classifier can induce a mismatch between the true label distributions of clean inputs and their adversarial examples.
    For simplicity we consider adversarial perturbation based on FGSM and cross-entropy loss, namely $x' = x -\varepsilon \|\nabla~f(x)_y\|^{-1} \nabla~f(x)_y$.
    The distribution mismatch induced by such adversarial perturbation can be lower bounded.
    \begin{lemma}
    \label{theorem:distribution-mismatch-true-model}
    Assume $f(x)_y$ is $L$-locally Lipschitz around $x$ with bounded Hessian. Let $\sigma_m = \inf_{z \in \mathcal{B}_\varepsilon(x)} \sigma_{\min} (\nabla^2 f(z)_y) > 0$ and $\sigma_M = \sup_{z \in \mathcal{B}_\varepsilon(x)} \sigma_{\max} (\nabla^2 f(z)_y) > 0$.
    Here $\sigma_{\min}$ and $\sigma_{\max}$ denote the minimum and maximum eigenvalues of the Hessian, respectively.
    We then have
    \begin{equation}
        \|P(Y|x) - P(Y'|x')\|_{\text{TV}} \ge
        \frac{\varepsilon}{2} (1 - f(x)_y) \frac{\sigma_m}{L}  - \frac{\varepsilon^2}{4} \sigma_M,
    \end{equation}
    \end{lemma}

    One can find that the right-hand side is positive as long as the upper bound of the Hessian norm is not too large, which is reasonable as previous works have shown that small hessian norm is critical to both standard~\citep{Keskar2017OnLT} and robust generalization~\citep{MoosaviDezfooli2019RobustnessVC}.

    \smallsection{Assigned label distribution is unchanged} Despite the fact that the true label distribution is distorted by adversarial perturbation, we note that the assigned label distribution of adversarial examples is still the same as their clean counterparts.
    \begin{remark} 
    \label{remark:common-practice}
     In adversarial training, it is the common practice that directly copies the label of a clean input to its adversarial counterpart, namely $\tilde{y}' = \tilde{y}$ and $P(\tilde{Y}'|x') = P(\tilde{Y}|x)$.
    \end{remark}

    

   \smallsection{Distribution mismatch indicates label noise}
   We show that a mismatch between the true label distribution and the assigned label distribution in a training set will always indicate the existence of label noise.
    \begin{lemma}
    \label{theorem:implicit-label-noise}
        Given a training set $\mathcal{D} = \{(x_i, \tilde{y}_i)\}_{i\in[N]}$, the label noise is lower-bounded by the mismatch between the true label distribution and the assigned label distribution. Specifically, with probability $1 - \delta$, we have
        \begin{equation}
        p_e(\mathcal{D}) \ge \mathbb{E}_x \|P(\tilde{Y}| x) - P(Y | x)\|_{\text{TV}} -\sqrt{\frac{1}{2N}\log\frac{2}{\delta}}
        \end{equation}
    \end{lemma}

   \smallsection{Label noise implicitly exists in adversarial training} In the adversarially augmented training set $\mathcal{D}'$, such distribution mismatch exists exactly. By Remark~\ref{remark:common-practice} we have $P(\tilde{Y}' | x') = P(\tilde{Y}|x)$ and by property of the clean dataset~(Assumption~\ref{assumption:clean-dataset}) we have $P(\tilde{Y}|x) = P(Y|x)$, which together means $P(\tilde{Y}' | x') = P(Y|x)$. However, Lemma~\ref{theorem:distribution-mismatch-true-model} shows that $P(Y' | x') \ne P(Y | x)$, which implies that $P(\tilde{Y}' | x') \ne P(Y' | x')$. This indicates that label noise exists in the adversarially augmented training set.
   We now have the following theorem, which is our main result.
  
    \begin{theorem}
    \label{theo:main}
    Assume $f(x)_y$ is $L$-locally Lipschitz around $x$ with Hessian bounded below. 
    Instantiate the same notations as in Lemma~\ref{theorem:distribution-mismatch-true-model}. With probability $1 - \delta$, we have
    \begin{equation}
    \label{theo:label-noise-dependence}
        p_e(\mathcal{D}') \ge \frac{\varepsilon}{2} (1 - q(\mathcal{D})) \frac{\sigma_m}{L}  - \frac{\varepsilon^2}{4} \sigma_M -\sqrt{\frac{1}{2N}\log\frac{2}{\delta}}
    \end{equation}
    \end{theorem}

    
    The above results suggest that as long as a training set is augmented by adversarial perturbation, but with assigned labels unchanged, label noise emerges. We demonstrate this by showing that standard training on a fixed adversarially augmented training set can also produce overfitting. Specifically, for each example in a clean training set we apply adversarial perturbation generated by a adversarially trained classifier. We then fix such an augmented training set and conduct standard training on it. We experiment on CIFAR-10 with WRN-28-5. A training subset of size 5k is randomly sampled to speed up the training. More details about the experiment settings can be found in the appendix. Figure~\ref{fig:fixed-augment} shows that prominent overfitting (as well as epoch-wise double descent) can be observed when the perturbation radius is relatively large.
    
    On the other hand, if a training set is augmented by perturbation that will not distort the true label distribution, there will not be label noise. We demonstrate this by showing that standard training on a training set augmented with Gaussian noise will not induce overfitting. As shown in Figure~\ref{fig:double-descent-gaussian}, even with a extremely large radius of Gaussian perturbation, no overfitting is observed. This also demonstrates that input perturbation not necessarily leads to overfitting.
    
\begin{figure}
    \centering
    \begin{minipage}{0.45\textwidth}
      \centering
      \includegraphics[width=1.0\linewidth]{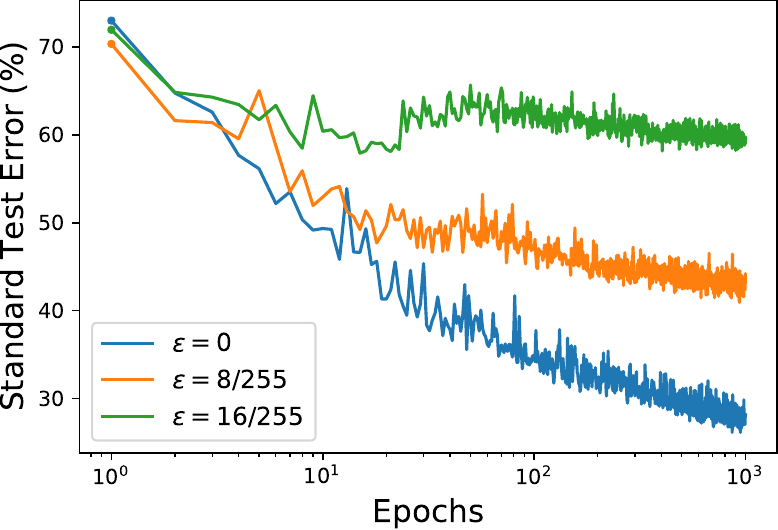}
      \caption{Standard training on a fixed adversarially augmented training set (e.g. $\varepsilon = 16/255$) can also produce prominent overfitting. In contrast, on the original training set without adversarial perturbation applied ($\varepsilon=0$), no overfitting is observed. 
      }
    \label{fig:fixed-augment}
    \vspace{-2mm}
    \end{minipage}\hfill
    \begin{minipage}{0.45\textwidth}
      \centering
      \includegraphics[width=1.0\textwidth]{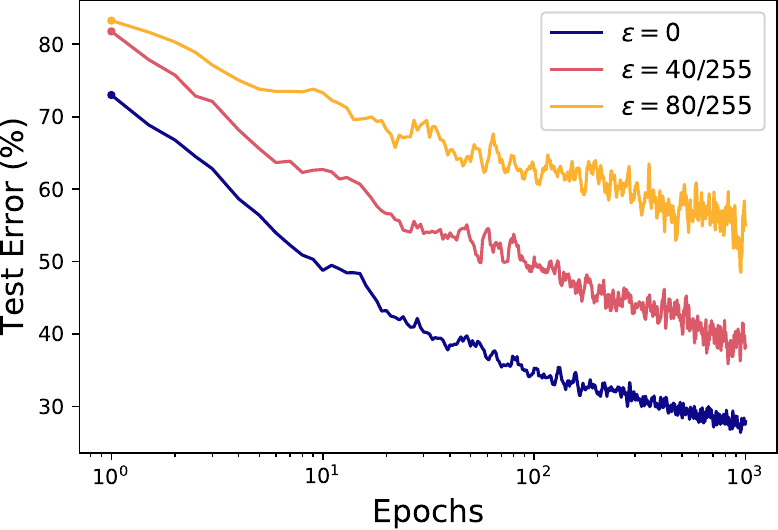}
      \caption{
      Standard training on a training set augmented by Gaussian noise will not produce overfitting. Here we select extremely large perturbation radius (e.g. $\varepsilon = 80/255$) to reduce the test error to be comparable to the adversarially augmented case. 
      }
    \label{fig:double-descent-gaussian}
    \end{minipage}
\end{figure}

\smallsection{Intuitive interpretation of label noise in adversarial training}
We introduce a simple example to help understand the emergence of label noise in adversarial training.
\begin{example}
[Label noise due to a symmetric distribution shift]
\label{example:label-noise-influence}
Let $\mathcal{D}=\{(x_i, y_i)\}_{i\in[N]}$ be a clean labeled training subset where all inputs $x_i=x$ are identical and have a one-hot true label distribution, i.e., $P(Y|x) = \mathbf{1}_y$. 

We now construct an adversarially augmented training subset $\mathcal{D'} = \{(x'_i, \tilde{y}'_i)\}_{i\in[N]}$, where $\tilde{y}' = y$ and $x'$ is generated based on adversarial perturbation that distorts the true label distribution symmetrically. Specifically,
$$
P(Y'= j' | x') =
\begin{cases} 
1 - \eta, & \text{if}~~j = y, \\
\eta/(K-1), & \text{otherwise}. \\
\end{cases}
$$
Then by Lemma~\ref{theorem:implicit-label-noise} 
we have $p_e (\mathcal{D}')\gtrsim \eta$.
\end{example}
   

One can find that there is indeed $\eta$ faction of noisy labels in $D'$. This is because if we sample the labels of $x'$ based on its true label distribution, we expect $1 - \eta$ faction of $x'$ are labeled as $y$, while $\eta$ fraction of $x'$ are labeled to be other classes. However, in $D'$, all $x'$ are assigned with label $y$ , which means $\eta$ fraction of $x'$ are labeled incorrectly. In realistic datasets we can consider inputs with similar features for such reasoning.



\begin{figure*}[!ht]
      \centering
      \includegraphics[width=0.8\textwidth]{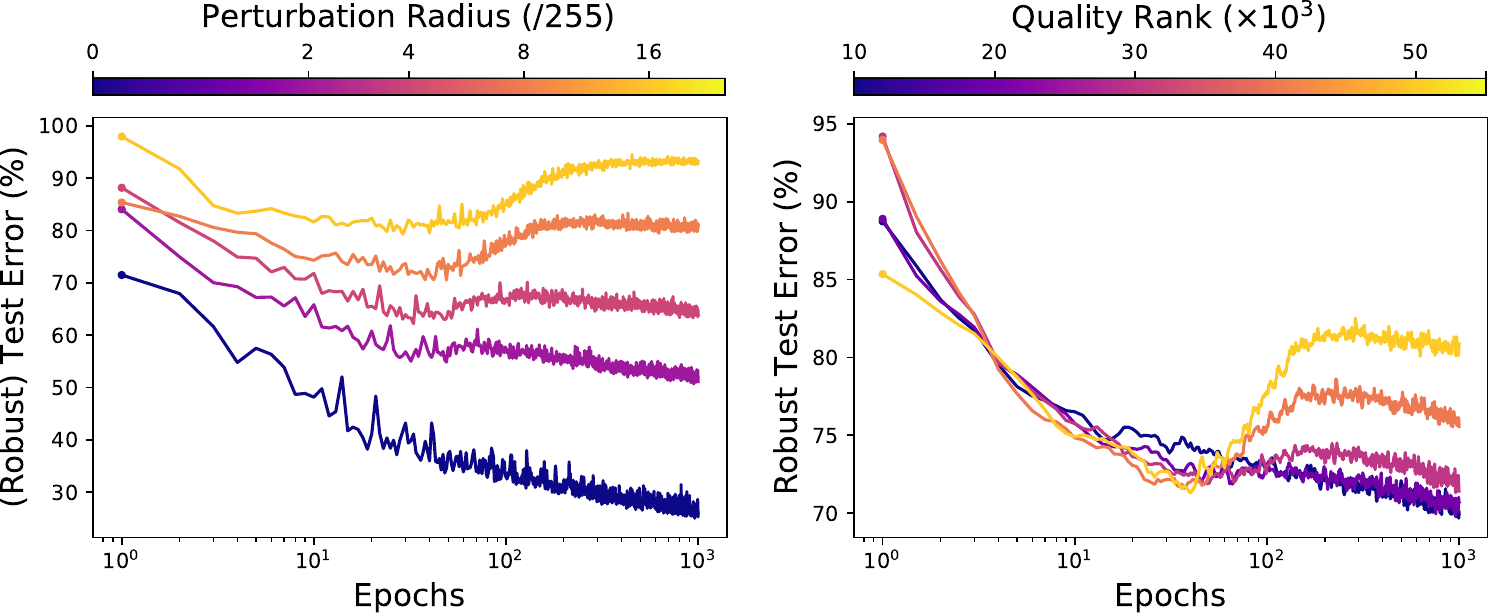}
      \caption{(Left) Dependence of robust overfitting on the perturbation radius. A training subset of size 5k is randomly sampled to speed up the training.
      $\varepsilon = 0/255$ indicates the standard training where no double descent occurs. 
      (Right) Dependence of robust overfitting on the data quality with a fixed perturbation radius ($\varepsilon = 8/255$). To construct a training subset with high data quality, we first calculate the predictive probability based on an ensemble of multiple models. We then rank all training examples based on the predictive probability and select the top-k ones.
      The curves are smoothed by a window of $5$ epochs to reduce overlapping.
      Here we conduct PGD training on CIFAR-10 with WRN-28-5. 
      More experiment details can be found in the Appendix. 
      }
      \vspace{-1em}
    \label{fig:dependence-perturbation-quality}
    \end{figure*}

The above example also shows that label noise in adversarial training may be stronger than one's impression. Even a slight distortion of the true label distribution, e.g. $\eta=0.1$, will be equivalent to at least $10\%$ noisy label in the training set. This is because the true label distribution of every training input is distorted, resulting in significant noise in the population. 
\smallsection{Dependence of label noise in adversarial training}
\label{sect:dependence-label-noise}
    Theorem~\ref{theo:main} shows that
    the label noise in adversarial training is proportional to (1) the perturbation radius (2) the data quality. 
    Considering label noise can be an important source of variance in the generalization of deep neural networks~\citep{Nakkiran2020DeepDD, Yang2020RethinkingBT}, such dependence of label noise explains the intriguing observations in the literature that robust overfitting (or epoch-wise double descent) in adversarial training will vanish with small perturbation radii~\citep{Dong2021ExploringMI} or high-quality data~\citep{Dong2021DataPF}. 
    We conduct more controlled experiments to verify this correlation empirically, as shown in Figure~\ref{fig:dependence-perturbation-quality}, 
    
\subsection{Adversarial perturbation generated by a realistic classifier}
\label{sect:reason-realistic}
We now consider a realistic case where the adversarial perturbation is generated by a probabilistic classifier $f_\theta$. 


\smallsection{Approximation of the true label distribution}
We show that after sufficient adversarial training, the predictive label distribution of $f_\theta$ can approximate the true label distribution with high probability. 
\begin{lemma}
\label{lemma:Learn-true-distribution}
Denote $\mathcal{S} = \{x: (x,y)\in \mathcal{D}\}$ as the collection of all training inputs.
Let $\rho\ge 1$ and $\mathcal{C}$ be an $\rho\varepsilon$-external covering of $\mathcal{S}$ with covering number $N_{\rho\varepsilon}$.
Let $f_\theta$ be a probabilistic classifier that minimizes the adversarial empirical risk (\ref{eq:outer-minimization}).
Assume $f_\theta$ is $L_\theta$-locally Lipschitz continuous in a norm ball of radius $\rho\varepsilon$ around $x \in \mathcal{C}$.
Let $\kappa \ge 1$ and $ \mathcal{\hat{S}}$ be a subset of $\mathcal{S}$ with cardinality at least $(1 - 1/\kappa + 1/(\kappa N_{\rho\varepsilon})) N$.
Let $\mathcal{N}_\varepsilon(\mathcal{\hat{S}})$ denote the neighborhood of the set $\mathcal{\hat{S}}$, i.e. $\mathcal{N}_\varepsilon(\mathcal{\hat{S}}) = \bigcup_{x\in\mathcal{\hat{S}}} \mathcal{B}_\varepsilon (x)$. 
Then for any $x \in \mathcal{N}_\varepsilon(\mathcal{\hat{S}})$, with probability at least $1-\delta$,
 
\begin{equation}
    \label{eq:main-theorem-bound}
    \|f_\theta(x) - P(Y|x)\|_{\text{TV}} \le \sqrt{\frac{\kappa N_{\rho\varepsilon} K}{2N}\log\frac{2}{\delta}}  + \left(\left(\frac{3}{2} - \frac{1}{K}\right) L_\theta + L\right) \rho \varepsilon,
\end{equation}


\end{lemma}


\smallsection{Label noise in adversarial training with a realistic classifier}
Adversarial perturbation generated by a realistic classifier $f_\theta$ will distort its predictive label distribution by gradient ascent. Subsequently, the true label distribution will also be distorted with high probability since the predictive label distribution of a realistic classifier $f_\theta$ can approximate the true label distribution. Specifically, by the triangle inequality we have
\begin{equation}
\|P(Y|x)  - P(Y'|x') \|_{\text{TV}} \ge \|f_\theta(x)  - f_\theta(x')\|_{\text{TV}} - (\|f_\theta(x) - P(Y|x)\|_{\text{TV}} + \|f_\theta(x') - P(Y'|x')\|_{\text{TV}}),
\end{equation}
where the last two terms are the approximation error of true label distribution on both clean and adversarial examples, which are guaranteed to be small. To conclude, we have the following result.
\begin{theorem}
\label{theo:realistic-classifier}
Instantiate the notations of Lemma~\ref{lemma:Learn-true-distribution}. For any $x \in \mathcal{N}_\varepsilon(\mathcal{\hat{S}})$, with probability at least $1 - 3\delta$, we have
\begin{equation}
    p_e(\mathcal{D}') \ge
    \varepsilon \left[(1 - \mathbb{E}_x f_\theta(x)_y) \frac{\sigma_m}{2L_\theta}- 2\rho\left(\left(\frac{3}{2} - \frac{1}{K}\right) L_\theta + L\right)\right]  - \varepsilon^2\frac{\sigma_M}{4}  
    - \xi \sqrt{\frac{1}{2N} \log\frac{2}{\delta}},
\end{equation}
where $\xi = 1 + \sqrt{4\kappa N_{\rho\varepsilon} K}$.
\end{theorem}

\section{Mitigate Label Noise in Adversarial Training}
\label{sect:mitigate-double-descent}

Since the label noise is incurred by the mismatch between the true label distribution and assigned label distribution of adversarial examples in the training set, we wish to find an alternative label (distribution) for the adversarial example to reduce such distribution mismatch.
We've already shown that the predictive label distribution of a classifier trained by conventional adversarial training, which we denote as \emph{model probability} in the following discussion, can in fact approximate the true label distribution. 
Here we show that it is possible to further improve the predictive label distribution and reduce the label noise by calibration.
    

\subsection{Rectify model probability to reduce distribution mismatch}
\label{sect:approximate-true-distribution}

We show that it is possible to reduce the distribution mismatch by \emph{temperature scaling}~\citep{Hinton2015DistillingTK, Guo2017OnCO} enabled in the softmax function.
\begin{theorem}[Temperature scaling can reduce the distribution mismatch]
\label{theorem: model-probability}
    Let $f_\theta(x; T)$ denote the predictive probability of a probabilistic classifier scaled by temperature $T$, namely $f_\theta(x; T)_j = \exp(z_j/T) / (\sum_j \exp(z_j / T)), $
    where $z$ is the logits of the classifier from $x$. 
    Let $x'$ be an adversarial example correctly classified by a classifier $f_\theta$, i.e. $\argmax_j f_\theta(x')_j = y'$,
    then there exists $T$, such that
    $$
    \| f_\theta(x'; T) - P(Y'|x') \|_{TV} \le \| f_\theta(x') - P(Y' | x')\|_{TV}.
    $$

\end{theorem}


    Another way to further reduce the distribution mismatch is to interpolate between the model probability and the one-hot assigned label.
    We show that the interpolation works specifically for incorrectly classified examples and thus can be viewed as a complement to temperature scaling.
    \begin{theorem}[Interpolation can further reduce the distribution mismatch]
    \label{theorem: model-probability-coefficient}
        Let $x'$ be an adversarial example incorrectly classified by a classifier $f_\theta$, i.e. $\argmax_j f_\theta(x'; T)_j \ne y'$. 
        Assume $\max_j P(Y'=j|x') \ge 1/2$, then there exists an interpolation ratio $\lambda$, such that
        \begin{equation*}
        \small
         \| f_\theta(x'; T, \lambda) - P(Y' | x') \|_{TV} \le \| f_\theta(x'; T) - P(Y' | x')\|_{\text{TV}},
        \normalsize
        \end{equation*}
        where $f_\theta(x'; T, \lambda) = \lambda \cdot f_\theta(x'; T) + (1 - \lambda)\cdot P(\tilde{Y}' | x')$.
    \end{theorem}

    As a summarization, 
    to reduce the distribution mismatch, we propose to use 
    $f_\theta(x'; T, \lambda)$
    as the assigned label of the adversarial example in adversarial training, which we refer as the \emph{rectified model probability}.
    
    In Appendix~\ref{sect:optimal-temperature-mixup}, we show that the optimal hyper-parameters (i.e. $T$ and $\lambda$) of almost all training examples concentrate on the same set of values by studying on a synthetic dataset with known true label distribution. 
    Therefore it is possible to find an universal set of hyper-parameters that reduce the distribution mismatch for all adversarial examples. 

\subsection{Determine the optimal temperature and interpolation ratio}

The set of temperature and interpolation ratio in the rectified model probability that maximally reduces the distribution mismatch is not straightforward to find 
as the true label distribution of the adversarial example is unknown in reality. Fortunately, given a sufficiently large validation dataset as a whole, it is possible to measure the overall distribution mismatch in a frequentist's view without knowing the true label distribution of every single example. A popular metric adopted here is the negative log-likelihood (NLL) loss, which is known as a proper scoring rule~\citep{Gneiting2007StrictlyPS} and is also employed in the confidence calibration of deep networks~\citep{Guo2017OnCO}. By Gibbs's inequality it is easy to show that the NLL loss will only be minimized when the assigned label distribution matches the true label distribution~\citep{Hastie2001TheEO}, namely


\begin{equation}
- \mathbbm{E}_{(x', y')\in \mathcal{D}'_\text{val}} \log f_\theta(x'; T, \lambda)_{y'}
\ge - \mathbbm{E}_{P(Y)} P(Y'|x') \log P(Y'|x').
\end{equation}

Therefore, we propose to find the optimal $T$ and $\lambda$ 
as
\begin{equation}
    \label{eq:calibration}
    T, \lambda = \argmin_{T, \lambda} - \mathbbm{E}_{(x', y')\in \mathcal{D}'_\text{val}} \log f_\theta(x'; T, \lambda)_{y'}.
\end{equation}

\begin{wrapfigure}{r}{7cm} 
  \includegraphics[width=1.0\linewidth]{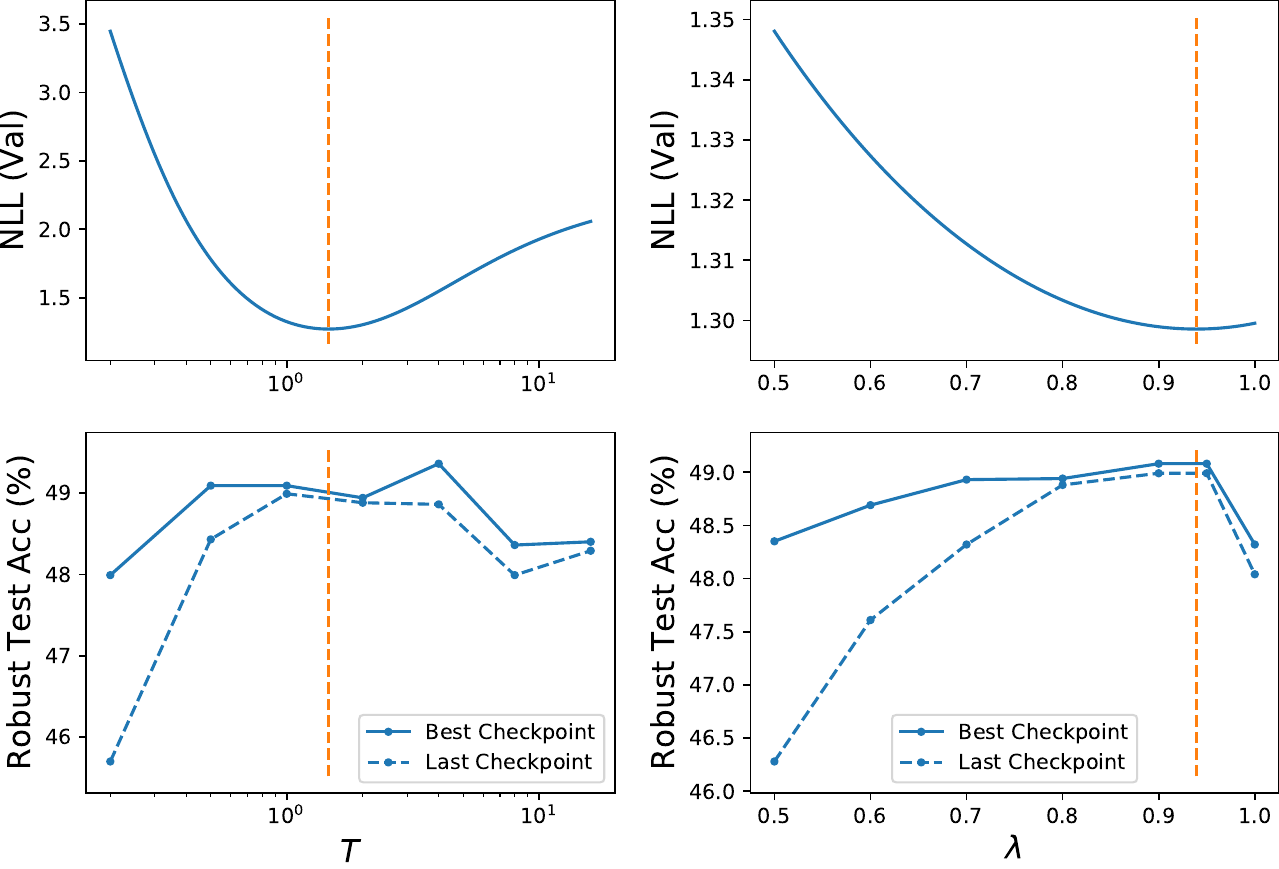}
  \caption{(Upper) NLL loss obtained on the validation set for different $T$ and $\lambda$. (Bottom) Robust test accuracy at the best and last checkpoint by adversarial training with the rectified model probability with different $T$ and $\lambda$. $\lambda=0.8$ for grid search on $T$ (Left) and $T=2$ for grid search on $\lambda$ (Right). Orange dashed lines indicate $T$ and $\lambda$  determined by Equation~(\ref{eq:calibration}).}%
  \label{fig:method-grid-search}
  \vspace{-2cm}
\end{wrapfigure}

\subsection{Rectified model probability mitigates robust overfitting}
\label{sect:exp-practical-adversarial-training}


We now work on a realistic dataset (CIFAR-10) to demonstrate the rectified model probability 
can effectively mitigate the robust overfitting, or equivalently the epoch-wise double descent in adversarial training. The outer minimization of adversarial training (Equation~(\ref{eq:outer-minimization})) now becomes 
    \begin{equation}
        \theta^* = \argmin_\theta \mathbbm{E}_\mathcal{D'}~ \ell\left(f_\theta(x'), f_{\hat{\theta}}(x'; T, \lambda)_{y'}\right),
    \end{equation}
    where $\hat{\theta}$ denotes the parameters of a classifier adversarially trained beforehand.
    The details of the experimental setting are available in the Appendix. 


As shown in Figure~\ref{fig:method-grid-search}, adversarial training on rectified model probability can mitigate the robust overfitting when the temperature $T$ and interpolation ratio $\lambda$ are optimal. Such optimal hyperparameters perfectly aligns with the ones automatically determined by Equation~(\ref{eq:calibration}).

\section{Experiments}
\label{sect:experiment}

\FloatBarrier

\begin{figure*}[t]
  \centering
  \includegraphics[width=0.98\linewidth]{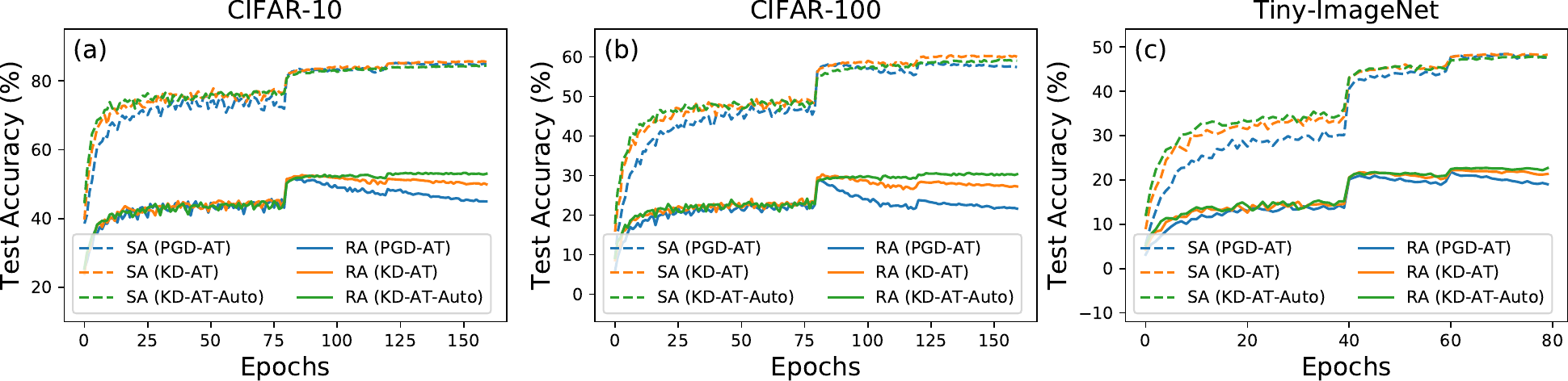}
  \vspace{-1ex}
  \caption{Our method can effectively mitigate robust overfitting for different datasets. 
  }
\label{fig:mitigate-overfitting}
\end{figure*}


\smallsection{Experiment setup}
We conduct experiments on three datasets including CIFAR-10, CIFAR-100~\citep{Krizhevsky2009LearningML} and Tiny-ImageNet~\citep{Le2015TinyIV}. We conduct PGD training on pre-activation ResNet-18~\citep{He2016IdentityMI} with $10$ iterations and perturbation radius $8/255$ by default. We evaluate robustness against $\ell_\infty$ norm-bounded adversarial attack with perturbation radius $8/255$, and employ AutoAttack~\citep{Croce2020ReliableEO} for reliable evaluation.  
Appendix~\ref{sect:more-result} includes results on additional model architectures (e.g., VGG~\citep{Simonyan2015VeryDC}, WRN),  adversarial training methods (e.g., TRADES~\citep{Zhang2019TheoreticallyPT}, FGSM~\citep{Goodfellow2015ExplainingAH}), and evaluation metrics (e.g., PGD-1000 (PGD attack with $1000$ iterations), Square Attack~\citep{Andriushchenko2020SquareAA}, RayS~\citep{Chen2020RaySAR}). More setup details can be found in Appendix~\ref{sect:exp-setting-all}.


\begin{table*}[!t]
  \small
  \caption{Performance of our method on different datasets. $^*$ denotes the hyper-parameters automatically determined by our method. 
   }
  \vspace{0.5ex}
  \label{table:result-dataset}
  \centering
  \small
  \begin{tabular}{clcccccccc}
    \toprule
    \multirow{2}{*}{Dataset} & \multirow{2}{*}{Setting} & \multirow{2}{*}{$T$} & \multirow{2}{*}{$\lambda$} & \multicolumn{3}{c}{Robust Acc. (\%)} & \multicolumn{3}{c}{Standard Acc. (\%)}\\
     & & &  & Best & Last & Diff. & Best & Last & Diff.\\
    \midrule
\multirow{3}{*}{CIFAR-10} 
& AT & - & - &  $47.35$ & $41.42$ & $ 5.93$ &  $82.67$ &  $84.91$ & -$2.24$ \\
 & KD-AT & $2$ & $0.5$ &  $48.76$ & $46.33$ & $ 2.43$ &  $82.89$ &  $\textbf{85.49}$ & -$2.60$ \\ 
 & KD-AT-Auto & $1.47^*$ & $0.8^*$ &  $\textbf{49.05}$ & $\textbf{48.80}$ & $ \textbf{0.25}$ &  $\textbf{84.26}$ &  $84.47$ & $\textbf{-0.21}$ \\ 
    \midrule
\multirow{3}{*}{CIFAR-100}
& AT & - & - &  $24.79$ & $19.75$ & $ 5.04$ &  $57.33$ &  $57.42$ & -$0.09$ \\ 
& KD-AT & $2$ & $0.5$ &  $25.77$ & $23.58$ & $ 2.19$ &  $57.24$ &  $\textbf{60.04}$ & -$2.80$ \\ 
& KD-AT-Auto & $1.53^*$ & $0.83^*$ &  $\textbf{26.36}$ & $\textbf{26.24}$ & $\textbf{0.12}$ &  $\textbf{58.80}$ &  $59.05$ & $\textbf{-0.25}$ \\ 
\midrule
\multirow{3}{*}{Tiny-ImageNet} 
& AT & - & - &  $17.20$ & $15.40$ & $ 1.80$ &  $47.72$ &  $47.62$ & $ 0.10$ \\ 
& KD-AT & $2$ & $0.5$ &  $17.86$ & $17.18$ & $ 0.68$ &  $\textbf{47.73}$ &  $\textbf{48.28}$ & -$0.55$ \\ 
& KD-AT-Auto  & $1.23^*$ & $0.85^*$ &  $\textbf{18.29}$ & $\textbf{18.39}$ & $\textbf{-0.10}$ &  $47.46$ &  $47.56$ & $\textbf{-0.10}$ \\ 
    \bottomrule
  \end{tabular}
\end{table*}

\smallsection{Results \& Discussions}
Our method is essentially the baseline adversarial training with a robust-trained self-teacher, equipped with an algorithm automatically deciding the optimal hyper-parameters, which we now denote as KD-AT-Auto.
We compare KD-AT-Auto with two baselines: regular adversarial training (AT), and adversarial training combined with self-distillation (KD-AT) with fixed temperature $T=2$ and interpolation ratio $\lambda=0.5$ as suggested by \citet{chen2021robust}. 

As shown in Figure~\ref{fig:mitigate-overfitting}, our method can effectively mitigate robust overfitting for all datasets, with both standard accuracy (SA) and robust accuracy (RA) constantly increasing throughout training. In Table~\ref{table:result-dataset}, we measure the difference between the RA at the best checkpoint (Best) and at the last checkpoint (Last) to clearly show the overfitting gap. Our method can reduce the overfitting gap to less than $0.5\%$ for all datasets. One may note that self-distillation with fixed hyper-parameters is in fact inferior in terms of reducing robust overfitting, while its effectiveness can be significantly improved with the optimal hyper-parameters automatically determined by our method, which further verifies our understanding of robust overfitting. 
Compared with self-distillation with fixed hyper-parameters, our method can also boost both RA and SA at the best checkpoint for all datasets.

Our method can further be combined with orthogonal techniques such as Stochastic Weight Averaging (SWA)~\citep{Izmailov2018AveragingWL} and additional standard teachers as mentioned in previous work~\citep{chen2021robust} to achieve better performance. More results and discussion can be found in Appendix~\ref{sect:additional-technique}.

\section{Conclusion and Discussions}
\label{sect:conclusion}


In this paper, we show that label noise exists implicitly in adversarial training due to the mismatch between the true label distribution and the assigned label distribution of adversarial examples. Such label noise can explain the dominant overfitting phenomenon. Based on a label noise perspective, we also extend the understanding of robust overfitting and show that it is the early part of an epoch-wise double descent in adversarial training. Finally, we propose an alternative labeling of adversarial examples by rectifying model probability, which can effectively mitigate robust overfitting without any manual hyper-parameter tuning. 

The label noise implicitly exists in adversarial training may have other important effects on adversarially robust learning. This can potentially consolidate the theoretical grounding of robust learning. For instance, since label noise induces model variance, from a model-wise view, one may need to increase model capacity to reduce the variance. This may partially explain why robust generalization requires significantly larger model than standard generalization. 
\bibliography{icml2022}
\bibliographystyle{icml2022}

\section*{Checklist}

The checklist follows the references.  Please
read the checklist guidelines carefully for information on how to answer these
questions.  For each question, change the default \answerTODO{} to \answerYes{},
\answerNo{}, or \answerNA{}.  You are strongly encouraged to include a {\bf
justification to your answer}, either by referencing the appropriate section of
your paper or providing a brief inline description.  For example:
\begin{itemize}
  \item Did you include the license to the code and datasets? \answerYes{See Section~\ref{gen_inst}.}
  \item Did you include the license to the code and datasets? \answerNo{The code and the data are proprietary.}
  \item Did you include the license to the code and datasets? \answerNA{}
\end{itemize}
Please do not modify the questions and only use the provided macros for your
answers.  Note that the Checklist section does not count towards the page
limit.  In your paper, please delete this instructions block and only keep the
Checklist section heading above along with the questions/answers below.

\begin{enumerate}

\item For all authors...
\begin{enumerate}
  \item Do the main claims made in the abstract and introduction accurately reflect the paper's contributions and scope?
    \answerYes{}
  \item Did you describe the limitations of your work?
    \answerYes{See Section~\ref{sect:limitation}}
  \item Did you discuss any potential negative societal impacts of your work?
    \answerYes{See Section~\ref{sect:limitation}}
  \item Have you read the ethics review guidelines and ensured that your paper conforms to them?
    \answerYes{}
\end{enumerate}

\item If you are including theoretical results...
\begin{enumerate}
  \item Did you state the full set of assumptions of all theoretical results?
    \answerYes{}
        \item Did you include complete proofs of all theoretical results?
    \answerYes{}
\end{enumerate}

\item If you ran experiments...
\begin{enumerate}
  \item Did you include the code, data, and instructions needed to reproduce the main experimental results (either in the supplemental material or as a URL)?
    \answerYes{}
  \item Did you specify all the training details (e.g., data splits, hyperparameters, how they were chosen)?
    \answerYes{}
        \item Did you report error bars (e.g., with respect to the random seed after running experiments multiple times)?
    \answerYes{}
        \item Did you include the total amount of compute and the type of resources used (e.g., type of GPUs, internal cluster, or cloud provider)?
    \answerYes{}
\end{enumerate}

\item If you are using existing assets (e.g., code, data, models) or curating/releasing new assets...
\begin{enumerate}
  \item If your work uses existing assets, did you cite the creators?
    \answerYes{}
  \item Did you mention the license of the assets?
    \answerNA{}
  \item Did you include any new assets either in the supplemental material or as a URL?
    \answerNA{}
  \item Did you discuss whether and how consent was obtained from people whose data you're using/curating?
    \answerNA{}
  \item Did you discuss whether the data you are using/curating contains personally identifiable information or offensive content?
    \answerNA{}
\end{enumerate}

\item If you used crowdsourcing or conducted research with human subjects...
\begin{enumerate}
  \item Did you include the full text of instructions given to participants and screenshots, if applicable?
    \answerNA{}
  \item Did you describe any potential participant risks, with links to Institutional Review Board (IRB) approvals, if applicable?
    \answerNA{}
  \item Did you include the estimated hourly wage paid to participants and the total amount spent on participant compensation?
    \answerNA{}
\end{enumerate}

\end{enumerate}


\newpage
\appendix\onecolumn
\section{Proofs}
\label{sect:proof-all}

\subsection{Preliminaries}
\smallsection{Proof in Assumption~\ref{assumption:clean-dataset}}
Here we prove that if there no label error in the clean dataset, then $P(\tilde{Y}|x) = P(Y|x)$.

\begin{proof}
First, we note that 
$$
    P(\tilde{Y}=j'|x) = \sum_j P(\tilde{Y}=j'| Y=j, x) P(Y=j|x).
$$
Since $P(E=1| Y=j, x) = 0$ we have,
$$
    \begin{aligned}
    P(\tilde{Y}=j'| Y=j, x)
    & = \sum_e P(\tilde{Y}=j'| E=e, Y=j, x) P(E=e| Y=j, x)  \\
    & = P (\tilde{Y}=j'| E=0, Y=j, x) \\
    & = 
    \begin{cases}
    1, & ~\text{if}~j'=j,\\
    0, & ~\text{otherwise}.
    \end{cases}
    \end{aligned}
$$
Therefore we have for all $j'$,
$$
    P(\tilde{Y}=j'|x) = P(Y=j'|x). 
$$
\end{proof}

\smallsection{Total Variation distance for discrete probability distributions}
    For two discrete probability distributions $P(Y)$ and $P(Y')$ where $Y, Y'\in\mathcal{Y}$, the total variation distance between them can be equally defined as
    $$
     \begin{aligned}
       \|P(Y) - P(Y')\|_{\text{TV}} 
       & = \sup_{J\in \mathcal{A}} \left| P(Y\in J) - P(Y'\in J)\right| \\
       & = \sup_{J\in \mathcal{A}} \left| \sum_{j\in J} P(Y=j) - \sum_{j\in J}P(Y'=j)\right| \\
      & = \frac{1}{2}\sum_j |P(Y=j) - P(Y'=j|)| \\
     \end{aligned}
    $$

\subsection{Proofs in Section~\ref{sect:reason-true}}
\label{sect:label-noise-more-proof}

\smallsection{Proof of Lemma~\ref{theorem:distribution-mismatch-true-model}}

\begin{proof}
For simplicity, we consider the adversarial perturbation generated by FGSM. Other adversarial perturbation can be viewed as a Taylor series of such perturbation.

    \begin{equation}
        \delta = -\varepsilon  \frac{\nabla~f(x)_y}{\|\nabla~ f(x)_y\|},  
    \end{equation}

First, we bound the distribution mismatch by gradient norm.
$$
    \begin{aligned}
    \|P(Y|x) - P(Y'|x')\|_{\text{TV}}
    & = \frac{1}{2} \sum_j \left|P(Y=j|x) - P(Y'=j|x')\right|\quad \boxed{\text{TV distance}}\\
    & \ge \frac{1}{2} \left|P(Y=y|x) - P(Y'=y|x')\right|\\
    & = \frac{1}{2} \left|f(x)_{y} - f(x')_{y}\right| \\
    & = \frac{1}{2} \left[ -\nabla f(x)_{y} \cdot \delta - \frac{1}{2}\delta^T \nabla^2 f(z)_{y} \delta \right]\\
    & \ge \frac{1}{2} \left[-\nabla f(x)_{y} \cdot \delta - \frac{\sigma_M}{2}\|\delta\|_2^2\right]  \quad \boxed{\text{Bounded Hessian}} \\ 
    & \ge \frac{1}{2} \left[\varepsilon \frac{\|\nabla f(x)_{y}\|^2_2}{\|\nabla f(x)_{y}\|} - \frac{\sigma_M}{2} \varepsilon^2 \frac{\|\nabla f(x)_{y}\|^2_2}{\|\nabla f(x)_{y}\|^2}  \right].\\
    \end{aligned}
$$
Now if $\|\cdot\| = \|\cdot\|_2$, we have
\begin{equation}
  \label{eq:mismatch-bound-l2}
  \|P(Y|x) - P(Y'|x')\|_{\text{TV}} \ge \frac{1}{2} \left[\varepsilon \|\nabla f(x)_{y}\|_2 - \frac{\sigma_M}{2}\varepsilon^2\right].    
\end{equation}
If $\|\cdot\| = \|\cdot\|_\infty$, we can utilize the fact that $\|\cdot\|_\infty \le \|\cdot\|_2 \le \sqrt{d} \|\cdot\|_\infty$,
thus 
\begin{equation}
    \label{eq:mismatch-bound-linf}
    \|P(Y|x) - P(Y'|x')\|_{\text{TV}} \ge \frac{1}{2} \left[\varepsilon \|\nabla f(x)_{y}\|_\infty - \frac{\sigma_M}{2} \varepsilon^2 \sqrt{d}\right].
\end{equation}
    
Second, we bound the gradient norm by the $L$-local Lipschitzness assumption.
    Let $x^*$ be a closest input that achieves the local maximum on the predicted probability at $y$, namely $x^* = \argmin_{z \in X, f(z)_y = 1} \|x - z\|$. Because $x^*$ is the local maximum and $f$ is continuously differentiable, $\nabla f(x^*)_y = 0$, thus
    $$
    \nabla f(x)_y 
    = \nabla f(x^*)_y  + \nabla^2 f(z)_y  (x - x^*) = \nabla^2 f(z)_y  (x - x^*).
    $$
    Therefore we have
    $$
    \begin{aligned}
    \|\nabla f(x)_y \| 
    & =  \|\nabla^2 f(z)_y  (x - x^*) \| \\
    & \ge \sigma_{m} \|x - x^*\| \\
    & \ge \sigma_{m}  \frac{|f(x^*)_y  - f(x)_y |}{L} \\
    & = \frac{\sigma_{m}}{L} (1 - f(x)_y). \\
    \end{aligned}
    $$
Plug this into Equation~(\ref{eq:mismatch-bound-l2}) or Equation~(\ref{eq:mismatch-bound-linf}) we then obtain the desired result.
\end{proof}

\smallsection{Proof of Lemma~\ref{theorem:implicit-label-noise}}
    \begin{proof}
    
    First, we show that the expectation of the label error is lower bounded by the mismatch between the true label distribution and the assigned label distribution.
    \begin{equation}
    \begin{aligned}
    \|P(\tilde{Y}|x) - P(Y|x)\|_{TV} 
    & = \frac{1}{2} \sum_j |P(\tilde{Y}=j|x) - P(Y=j|x)|  \\
    & = \frac{1}{2} \sum_j |P(\tilde{Y}=j, Y=j|x) + P(\tilde{Y}=j, Y\ne j|x) \\
    & \quad - P(Y=j, \tilde{Y}=j|x)- P(Y=j, \tilde{Y}\ne j|x)| \\
    & =  \frac{1}{2} \sum_j | P(\tilde{Y}=j, Y\ne j|x) - P(Y=j, \tilde{Y}\ne j|x)| \\
    & \le \frac{1}{2} \sum_j  P(\tilde{Y}=j, Y\ne j|x) + P(Y=j, \tilde{Y}\ne j|x) \\
    & = P(Y’\ne Y | x) \\
    & = P(E=1|x)\\
    \end{aligned}
    \end{equation}
    
    Second, given a sampled training set $\mathcal{D}=\{(x_i, \tilde{y}_i)\}_{i\in [N]}$, the empirical measure of label error $E$ should converge to its expectation almost surely, namely
    $$
    \lim_{N\to \infty} p_e(\mathcal{D}) = \lim_{N\to \infty} \frac{1}{N} \sum_{i\in [N]} e_i = \mathbb{E} [E] = P(E = 1).
    $$
    Using standard concentration inequality such as Hoeffding's inequality we have, 
    with probability $ 1 - \delta$,
    $$
    |p_e(\mathcal{D}) - P(E = 1)| \le \sqrt{\frac{1}{2N}\log\frac{2}{\delta}}.
    $$
    This implies
    $$
    p_e(\mathcal{D}) \ge P(E=1) -  \sqrt{\frac{1}{2N}\log\frac{2}{\delta}}.    
    $$
    
    Since $P(E = 1) = \mathbb{E}_x P(E=1|x)$, we have, with probability $1 - \delta$,
    $$
    p_e(\mathcal{D}) \ge \mathbb{E}_x \|P(\tilde{Y}| x) - P(Y | x)\|_{\text{TV}} -\sqrt{\frac{1}{2N}\log\frac{2}{\delta}} .
    $$
    which means $p_e(\mathcal{D}) > 0$ as long as $N$ is large.
    
    \end{proof}

\smallsection{Proof of Theorem~\ref{theo:main}}
    \begin{proof}
    First, by the fact that $P(\tilde{Y}'| x') = P(\tilde{Y} | x)$ and $P(\tilde{Y} | x) = P(Y | x)$ we have $P(\tilde{Y}'| x') =  P(Y | x)$.
    
    Therefore, apply Lemma~\ref{theorem:implicit-label-noise} to an adversarially augmented training set we have with probability $1 - \delta$,
    $$
    \begin{aligned}
    p_e(\mathcal{D'}) 
    & \ge \mathbb{E}_x \|P(\tilde{Y}'| x') - P(Y' | x')\|_{\text{TV}} -\sqrt{\frac{1}{2N}\log\frac{2}{\delta}} \\
    & \ge \mathbb{E}_x \|P(Y | x) - P(Y' | x')\|_{\text{TV}} -\sqrt{\frac{1}{2N}\log\frac{2}{\delta}}. \\
    \end{aligned}
    $$
    Further, apply Lemma~\ref{theorem:distribution-mismatch-true-model} and the definition of data quality, we have with probability $1 - \delta$,
    $$
    \begin{aligned}
    p_e(\mathcal{D'}) 
    & \ge \frac{\varepsilon}{2} (1 - \mathbb{E}_x f(x)_y) \frac{\sigma_m}{L}  - \frac{\varepsilon^2}{4} \sigma_M - \sqrt{\frac{1}{2N}\log\frac{2}{\delta}} \\
    & \ge \frac{\varepsilon}{2} (1 - q(\mathcal{D})) \frac{\sigma_m}{L}  - \frac{\varepsilon^2}{4} \sigma_M - \sqrt{\frac{1}{2N}\log\frac{2}{\delta}}. \\    
    \end{aligned}
    $$
    
    \end{proof}

\subsection{Proofs in Section~\ref{sect:reason-realistic}}
\label{sect:label-noise-proof-realistic}

\smallsection{Proof of Lemma~\ref{lemma:Learn-true-distribution}}    
\begin{proof}


Let $\mathcal{D} = (x, y)$ be the adversarially augmented training set. Let $\mathcal{S} = \{x: (x,y)\in \mathcal{D}\}$ be the collection of all training inputs.
First we note that the set of all training inputs can be grouped into several subsets such that the inputs in each subset possess similar true label distribution. 
More formally, 
Let $\mathcal{C} = \{\bar{x}_j\}_{j=1}^{N_{\rho\varepsilon}}$ be an $\rho\varepsilon$-external covering of $\mathcal{S}$ with minimum cardinality, namely $\mathcal{S} \subseteq \bigcup_{x \in \mathcal{C}} \{x' ~|~ \|x ' - x\| \le \rho\varepsilon \}$,
where we refer $\bar{x}_j$ as the covering center, and $N_{\rho\varepsilon}$ is the covering number.

Let $\{\mathcal{S}_j\}_{j=1}^{N_{\rho\varepsilon}}$ be any disjoint partition of $\mathcal{S}$ such that $\mathcal{S}_j \subseteq \{x' | \|x ' - \bar{x}_j\| \le \rho\varepsilon \}$. 
We show that $\mathcal{{S}}_j$ attains a property that the true label distribution of any input $x$ in this subset will not be too far from the sample mean of one-hot labels $\mathbf{\bar{y}}_j = {|\mathcal{{S}}_j|}^{-1} \sum_{x\in \mathcal{{S}}_j} \mathbf{1}_{y}$ in this subset. Specifically, let $p(x) = P(Y|x)$. We have with probability $1 - \delta$,
\begin{equation}
     \label{lemma:sample-mean-relax}
    \| p(x) - \bar{\mathbf{y}}_j \|_1 \le \sqrt{\frac{2K}{|\mathcal{{S}}_j|}\log\frac{2}{\delta}} + 2L\rho\varepsilon.
\end{equation}

To prove this property we first present two lemmas. 
\begin{lemma}[Lipschitz constraint of the true label distribution]
\label{lemma:Lipschitz-true-distribution}
Let $\mathcal{S}_j$ be a subset constructed above and $\mathbf{\bar{y}}_j = {|\mathcal{{S}}_j|}^{-1} \sum_{x\in \mathcal{{S}}_j} \mathbf{1}_{y}$. Then for any $x \in \mathcal{S}_j$ we have,
\begin{equation}
    \left\|p(x) - \mathbb{E}[\mathbf{\bar{y}}_j] \right\|_1 \le 2 L \rho\varepsilon.
\end{equation}
\end{lemma}
\begin{proof}
First, since $x \in \mathcal{S}_j$, we have $\|x - \bar{x}_j\|_1 \le \rho\varepsilon $, which implies $\|p(x) - p(\bar{x}_j)\|_1 \le L \rho\varepsilon$ by the locally Lipschitz continuity of $p$. Then for any $x, x' \in \mathcal{S}_j$, we will have $\|p(x) - p(x')\| \le 2L \rho\varepsilon$ by the triangle inequality. Let $N_{\mathcal{S}} = |\mathcal{S}_j|$. Therefore,
\begin{equation}
    \left\|p(x) - \frac{1}{N_{\mathcal{S}}} \sum_{x \in \mathcal{S}_j} p(x)\right\|_1 \le 2 \frac{N_{\mathcal{S}} - 1}{N_{\mathcal{S}}} L \rho\varepsilon \le 2 L \rho\varepsilon.
\end{equation}
Further, the linearity of the expectation implies
\begin{equation}
    \mathbb{E}[\mathbf{\bar{y}}] = {N_{\mathcal{S}}}^{-1} \sum_{x \in \mathcal{S}_j} \mathbb{E}[\mathbf{1}_{y(x)}] = {N_{\mathcal{S}}}^{-1} \sum_{x \in \mathcal{S}_j} p(x).
\end{equation}
Therefore $ \left\|p(x) - \mathbb{E}[\mathbf{\bar{y}}_j] \right\| \le 2 L \rho\varepsilon$.
\end{proof}

\begin{lemma}[Concentration inequality of the sample mean]
\label{lemma:sample-mean}
Let $\mathcal{S}$ be a set of $x$ with cardinality $N$.
Let $\mathbf{\bar{y}} = N^{-1} \sum_{x\in \mathcal{S}} \mathbf{1}_y$ be the sample mean.
Then for any $p$-norm $\|\cdot\|$ and any $\varepsilon > 0$,  we have with probability $1 - \delta$,
\begin{equation}
    \label{eq:concentration-sample-mean}
    \left\| \mathbf{\bar{y}} - \mathbb{E}\left[ \mathbf{\bar{y}} \right] \right\|_1 \le \sqrt{\frac{2K}{N}\log \frac{2}{\delta}}
\end{equation}
\end{lemma}

\begin{proof}
Note that $\mathbf{\bar{y}}$ obeys a multinomial distribution, \emph{i.e.} $\mathbf{\bar{y}} \sim N^{-1} multinomial(N, \mathbb{E}[\mathbf{\bar{y}}])$.
This lemma is thus the classic result on the concentration properties of multinomial distribution based on $\ell_1$ norm~\cite{Weissman2003InequalitiesFT, Qian2020ConcentrationIF}.
\end{proof}

One can see that Lemma~\ref{lemma:Lipschitz-true-distribution} bounds the difference between true label distribution of individual inputs and the mean true label distribution, while Lemma~\ref{lemma:sample-mean} bounds the difference between the sample mean and the mean true label distribution. Therefore the difference between the true label distribution and the sample mean is also bounded, since by the triangle inequality we have with probability $1 - \delta$,
\begin{equation}
    \begin{aligned}
    \|p(x) - \bar{\mathbf{y}} \| 
    & \le \|p(x) - \mathbb{E}[\bar{\mathbf{y}}] \| + \| \bar{\mathbf{y}} - \mathbb{E}[\bar{\mathbf{y}}] \| \\
    & \le \sqrt{\frac{2K}{N}\log \frac{2}{\delta}} + 2L\rho\varepsilon.
    \end{aligned}
\end{equation}

\vspace{1em}
We now show that given the locally Lipschitz constraint established in each disjoint partition we constructed above, the prediction given by the empirical risk minimizer will be close to the sample mean. As an example, we focus on the negative log-likelihood loss, namely $\ell(f_\theta(x), y) = - \mathbf{1}_y \cdot \log f_\theta (x) $. Other loss functions that are subject to the proper scoring rule can be investigated in a similar manner.
First, we regroup the sum in the empirical risk based on the partition constructed above, namely
\begin{equation}
    \hat{R}(f_\theta, \mathcal{S}) = \frac{1}{N_{\rho\varepsilon}}\sum_{j=1}^{N_{\rho\varepsilon}} \hat{R} (f_\theta, \mathcal{S}_j),
\end{equation}
where $\hat{R} (f_\theta, \mathcal{S}_j) = -{|\mathcal{S}_j|}^{-1} \sum_{i=1}^{|\mathcal{S}_j|} \mathbf{1}_{y_i} \cdot \log f_\theta(x_i)$ is the empirical risk in each partition. Since we are only concerned with the existence of a desired minimizer of the empirical risk, we can view $f_\theta$ as able to achieve any labeling of the training inputs that suffices the local Lipschitz constraint. 
Thus the empirical risk minimization is equivalent to the minimization of the empirical risk in each partition.
The problem can thus be defined as, for each $j = 1, \cdots, N_{\rho\varepsilon}$,
\begin{equation}
\label{eq:minimum-partition}
\begin{aligned}
& \min_{f_\theta} \hat{R} (f_\theta, \mathcal{S}_j)\\
& s.t.~~ \|f_\theta(x) - f_\theta(\bar{x}_j)\|_1 \le L\rho\varepsilon, 
~\forall~x \in \mathcal{S}_j,
\end{aligned}
\end{equation}
where the constraint is imposed by the locally-Lipschitz continuity of $f_\theta$. By the following lemma, we show that the minimizer of such problem is achieved only if $f_\theta(\bar{x}_j)$ is close to the sample mean.

\begin{lemma}
\label{lemma:vector-minimization}
Let $\bar{\mathbf{y}} ={|\mathcal{S}_j|}^{-1} \sum_{x \in \mathcal{S}_j} \mathbf{1}_{y}$.  The minimum of the problem (\ref{eq:minimum-partition}) is achieved only if $f_\theta(\bar{x}_j) = \mathbf{\bar{y}_j}(1 + KL_\theta \rho\varepsilon) - L_\theta\rho\varepsilon$.
\end{lemma}

\begin{proof}
We note that since the loss function we choose is strongly convex, to minimize the empirical risk, the prediction of any input $x$ must be as close to the one-hot labeling as possible. Therefore the problem (\ref{eq:minimum-partition}) can be formulated into a vector minimization where we can employ Karush–Kuhn–Tucker (KKT) theorem to find the necessary conditions of the minimizer. 

Let $\mathbf{p}_i := f_\theta(x_i)$ and $\tilde{\varepsilon} = L\rho\varepsilon$ for simplicity.
We rephrase the problem~(\ref{eq:minimum-partition}) as 

\begin{equation}
\label{eq:vector-minimization2}
\begin{aligned}
    \min_{\{\mathbf{p}_i\}_{i=1}^N} & - \frac{1}{N} \sum_{i} \mathbf{1}_{y_i} \cdot \log \mathbf{p}_i \\
    s.t.~~& \|\mathbf{p}_i - \mathbf{p}\|_1 \le \tilde{\varepsilon}, ~\sum_k \mathbf{p}_i^k = 1, ~\sum_k \mathbf{p}^k = 1, \mathbf{p}^k_i \ge 0, \mathbf{p}^k \ge 0. \\
\end{aligned}
\end{equation}

\smallsection{Case I}
We first discuss the case when $\mathbf{p}^k + \tilde{\varepsilon} < 1$ for all $k$. 
First, we observe that for any $\mathbf{p}$, the minimum of the above problem is achieved only if $\mathbf{p}_i^{y_i} = \mathbf{p}^{y_i} + \tilde{\varepsilon}$. Because by contradiction, if $\mathbf{p}_i^{y_i} < \mathbf{p}^{y_i} + \tilde{\varepsilon}$, we will have $-\log \mathbf{p}_i^{y_i} > -\log(\mathbf{p}^{y_i} + \tilde{\varepsilon}) $, and $\mathbf{p}^{y_i} + \tilde{\varepsilon}$ belongs to the feasible set, which means $\mathbf{p}_i^{y_i}$ does not attain the minimum.

The above problem can then be rephrased as
\begin{equation}
\begin{aligned}
    \min_{\mathbf{p}} - \frac{1}{N} \sum_{i} \log( {\mathbf{p}}^{y_i} + \tilde{\varepsilon}),~~ s.t.~~\sum_k \mathbf{p}^k = 1, \mathbf{p}^k \ge 0,
\end{aligned}
\end{equation}
where we have neglected the condition associated with $\mathbf{p}_i^{k\ne y_i}$, since they do not contribute to the objective, they can be chosen arbitrarily as long as the constraints are sufficed, and clearly the constraints are underdetermined.

Let $N_k = \sum_i 1(y_i = k)$, we have $\sum_i \log (\mathbf{p}^{y_i} + \tilde{\varepsilon}) = \sum_k N_k \log (\mathbf{p}^k + \tilde{\varepsilon})$. Therefore the above problem is equivalent to
\begin{equation}
\begin{aligned}
    \label{eq:convex-minimization}
    \min_{\mathbf{p}}- \sum_k \mathbf{\bar{y}}^k \log (\mathbf{p}^k + \tilde{\varepsilon}),~~ s.t.~~\sum_k \mathbf{p}^k = 1, \mathbf{p}^k \ge 0,
\end{aligned}
\end{equation}
where $\mathbf{\bar{y}} \equiv [N_1 / N, \cdots, N_k / N]^T $ is equal to the sample mean $N^{-1}\sum_i \mathbf{1}_{y_i}$. 

To solve the strongly convex minimization problem (\ref{eq:convex-minimization}) it is easy to employ KKT conditions to show that  
$$ \mathbf{p} = \mathbf{\bar{y}} (1 + K\tilde{\varepsilon}) - \tilde{\varepsilon}. 
$$

\smallsection{Case II} We now discuss the case when $\mathbf{\hat{p}}$ is the minimizer of (\ref{eq:vector-minimization2}) and there exists $k'$ such that $\mathbf{\hat{p}}^{k'} + \tilde{\varepsilon} \ge 1$. And $\mathbf{\hat{p}}\ne p$, where $p = \mathbf{\bar{y}}(1+K\tilde{\varepsilon}) - \tilde{\varepsilon}$ is the form of the minimizer in the previous case.

Considering a non-trivial case $\mathbf{p^*}^{k'} < 1 - \tilde{\varepsilon}$. Otherwise the true label distribution is already close to the one-hot labeling, which is the minimizer of the empirical risk. Therefore by $\sum_{k\ne k'} {p}^{k} > \tilde{\varepsilon}$ we have the condition
\begin{equation}
    \sum_{k\ne k'}\mathbf{\bar{y}}^k >  \frac{K\tilde{\varepsilon}}{1 + K\tilde{\varepsilon}}
\end{equation}

Now considering the minimization objective $R(p) = - {N}^{-1} \sum_i \mathbf{1}_{y_i} \cdot \log \mathbf{p}_i$. For all $i$ with $y_i = k'$, we must have $\mathbf{p}_i^{y_i} = 1$, otherwise the optimal cannot be attained by contradiction. Then the minimization problem can be rephrased as
\begin{equation}
    \min \sum_{k\ne k'} \mathbf{\bar{y}}^k \log (\mathbf{\hat{p}} + \tilde{\varepsilon}), ~s.t. \sum_{k'\ne k} \mathbf{\hat{p}}^{k} \ge \tilde{\varepsilon}, \mathbf{\hat{p}}^{k} \ge 0,
\end{equation}
where the first constraint is imposed by $\mathbf{\hat{p}}^{k'} \ge 1 - \tilde{\varepsilon}$.

Employ KKT conditions similarly we can have $\mathbf{\hat{p}}^k = \mathbf{\bar{y}}^k/\lambda - \tilde{\varepsilon}$ where $\lambda$ is a constant. By checking the constraint we can derive $\lambda \ge \sum_k \mathbf{\bar{y}}^k / (K\tilde{\varepsilon})$.

However, the minimization objective 
$$
\min_{\lambda} - \sum_{k\ne k'} \mathbf{\bar{y}}^k \log \frac{\mathbf{\bar{y}}^k}{\lambda},
$$
requires $\lambda$ to be minimized. Therefore $\lambda = \sum_{k\ne k'} \mathbf{\bar{y}}^k / (K\tilde{\varepsilon})$, which implies
\begin{equation}
\mathbf{\hat{p}}^k = K\tilde{\varepsilon} \frac{\mathbf{\bar{y}}^k}{\sum_{k\ne k'} \mathbf{\bar{y}}^k} - \tilde{\varepsilon}.
\end{equation}

Now since $\mathbf{\hat{p}} = \arg\min_p R(p)$ and $\mathbf{\hat{p}} \ne p$, we must have $R(\mathbf{\hat{p}}) < R(p)$. This means
\begin{equation}
 - \sum_{k\ne k'} \mathbf{\bar{y}}^k \log \frac{K\tilde{\varepsilon} \mathbf{\bar{y}}^k}{\sum_{k\ne k'} \mathbf{\bar{y}}^k} < - \sum_{k\ne k'} \mathbf{\bar{y}}^k \log [\mathbf{\bar{y}}^k(1+K\tilde{\varepsilon})],
\end{equation}
which is reduced to
\begin{equation}
\sum_{k\ne k'}\mathbf{\bar{y}}^{k} < \frac{K\tilde{\varepsilon}}{1+K\tilde{\varepsilon}}
\end{equation}
But this is contradict to our assumption.
\end{proof}

We are now be able to bound the difference between the predictions of the training inputs produced by the empirical risk minimizer and the sample mean in each $\mathcal{S}_j$. To see that we have for each $x \in \mathcal{S}_j$. 
\begin{equation}
\begin{aligned}
    \|f_\theta(x) - \mathbf{\bar{y}}_j \|_1 
    & \le \|f_\theta(x) - f_\theta(\bar{x}_j) \|_1 + \|f_\theta(\bar{x}_j) - \mathbf{\bar{y}}_j \|_1  \\
    & \le L\rho\varepsilon ( 1 + K\|\mathbf{\bar{y}}_j -  K^{-1}\mathbf{1}\|_1)\\
    & \le L\rho\varepsilon ( 1 + K\|\mathbf{1}_{(\cdot)} - K^{-1}\mathbf{1}\|_1).\\
    & = L\rho\varepsilon\left(3 - \frac{2}{K}\right)
\end{aligned}
\end{equation}
By Equation~(\ref{lemma:sample-mean-relax}) we then have for any $x \in \mathcal{S}_j$, with probability $1 - \delta$,
\begin{equation}
    \label{eq:final-bound-each-partition}
    \|f_\theta(x) - p(x)\|_1 \le \sqrt{\frac{2K}{|\mathcal{S}_j|}\log\frac{2}{\delta}} + L_\theta\rho\varepsilon\left(3 - \frac{2}{K}\right) + 2L\rho\varepsilon,\\ 
\end{equation}
which means the difference between the predictions and the true label distribution is also bounded.

\smallsection{Step III: Show the disjoint partition is non-trivial}
In (\ref{eq:final-bound-each-partition}), we have managed to bound the difference between the predictions yielded by an empirical risk minimizer and the true label distribution based on the cardinality of the subset $|\mathcal{S}_j|$, namely the number of inputs in $j$-partition. However $|\mathcal{S}_j|$ is critical to the bound here as if $|\mathcal{S}_j| = 1$, then (\ref{eq:final-bound-each-partition}) becomes a trivial bound. Here we show $|\mathcal{S}_j|$ is non-negligible based on simple combinatorics.

\begin{lemma}
\label{lemma:bound-partition}
Let $\{\mathcal{S}_j\}_{j=1}^{N_{\rho\varepsilon}}$ be a disjoint partition of the entire training set $\mathcal{S}$. Denote $\mathcal{S}(x)$ as the partition that includes $x$. Let $N(x) = |\mathcal{S}(x)|$ and $N = |\mathcal{S}|$.
 Then for any $\kappa \ge 1$,
 \begin{equation}
     \left| \left\{ x ~|~ N(x) \ge \frac{ N }{\kappa N_{\rho\varepsilon}}  \right\} \right| \ge \left(1 - \frac{1}{\kappa} + \frac{1}{\kappa N_{\rho\varepsilon}}\right)N.
 \end{equation}
\end{lemma}
\begin{proof}
We note that the problem is to show the minimum number of $x$ such that $N(x) \ge N / (\kappa N_{\rho\varepsilon})$. This is equivalent to find the maximum number of $x$ such that $N(x) \le N / (\kappa N_{\rho\varepsilon})$. Since we only have $N_{\rho\varepsilon}$ subsets, the maximum can be attained only if for $N_{\rho\varepsilon} - 1$ subsets $\mathcal{S}$, $|\mathcal{S}| = N / (\kappa N_{\rho\varepsilon})$. Otherwise, if for any one of these subsets $|\mathcal{S}| < N / (\kappa N_{\rho\varepsilon})$, then it is always feasible to let $|\mathcal{S}| = N / (\kappa N_{\rho\varepsilon})$ and the maximum increases. Similarly, if the number of such subsets is less than $N_{\rho\varepsilon} - 1$, then it is always feasible to let another subset subject to $|\mathcal{S}| = N / (\kappa N_{\rho\varepsilon})$ and the maximum increases. We can then conclude that at most $N(N_{\rho\varepsilon} - 1)/(\kappa N_{\rho\varepsilon})$ inputs can have the property $N(x) \le N / (\kappa N_{\rho\varepsilon})$.

\end{proof}
The above lemma basically implies when partitioning $N$ inputs into $N_{\rho\varepsilon}$ subsets, a large fraction of the inputs will be assigned to a subset with cardinality at least $N / (\kappa N_{\rho\varepsilon})$. Here $N_{\rho\varepsilon}$ is the covering number and is bounded above based on the property of the covering in the Euclidean space. Apply Lemma~\ref{lemma:bound-partition} to (\ref{eq:final-bound-each-partition}), and use the fact that $\|\cdot\|_{\text{TV}} = \|\cdot\|_1/2$ for category distributions, we then arrive at Lemma~\ref{lemma:Learn-true-distribution}.


\end{proof}

\smallsection{Proof of Theorem~\ref{theo:realistic-classifier}}
\begin{proof}
    First, we show that adversarial perturbation generated by a realistic classifier can change its predictive distribution. Considering adversarial perturbation based on FGSM and cross-entropy loss, namely $x' = x -\varepsilon \|\nabla~f_\theta(x)_y\|^{-1} \nabla~f_\theta(x)_y$, we can obtain a result similar to Lemma~\ref{theorem:distribution-mismatch-true-model}.
    \begin{lemma}
    \label{theorem:distribution-mismatch-real-model}
    Assume $f_\theta(x)_y$ is $L_\theta$-locally Lipschitz around $x$ with bounded Hessian. Let $\sigma_m = \inf_{z \in \mathcal{B}_\varepsilon(x)} \sigma_{\min} (\nabla^2 f_\theta(z)_y) > 0$ and $\sigma_M = \sup_{z \in \mathcal{B}_\varepsilon(x)} \sigma_{\max} (\nabla^2 f_\theta(z)_y) > 0$.
    Here $\sigma_{\min}$ and $\sigma_{\max}$ denote the minimum and maximum eigenvalues of the Hessian, respectively.
    We then have
    \begin{equation}
        \| f_\theta(x) -  f_\theta(x')\|_{\text{TV}} \ge
        \frac{\varepsilon}{2} (1 - f_\theta(x)_y) \frac{\sigma_m}{L_\theta}  - \frac{\varepsilon^2}{4} \sigma_M,
    \end{equation}
    \end{lemma}

Second, We prove that the true label distribution will be distorted by the adversarial perturbation generated by a realistic classifier. This is guaranteed if the predictive distribution of a realistic classifier can approximate the true label distribution. Specifically, by utilizing Lemma~\ref{theorem:distribution-mismatch-real-model} and Lemma~\ref{lemma:Learn-true-distribution}, we have with probability $1 - 2\delta$,
\begin{equation}
\begin{aligned}
& \|P(Y|x)  - P(Y'|x') \|_{\text{TV}} \\ 
\ge & \|f_\theta(x)  - f_\theta(x')\|_{\text{TV}} - (\|f_\theta(x) - P(Y|x)\|_{\text{TV}} + \|f_\theta(x') - P(Y'|x')\|_{\text{TV}})\\
\ge & \frac{\varepsilon}{2} (1 - f_\theta(x)_y) \frac{\sigma_m}{L_\theta}  - \frac{\varepsilon^2}{4} \sigma_M - \sqrt{\frac{2\kappa N_{\rho\varepsilon} K}{N}\log\frac{2}{\delta}}  - \left(\left(\frac{3}{2} - \frac{1}{K}\right) L_\theta + L\right)2\rho\varepsilon \\
= & \varepsilon \left[(1 - f_\theta(x)_y) \frac{\sigma_m}{2L_\theta} - 2\rho\left(\left(\frac{3}{2} - \frac{1}{K}\right) L_\theta + L\right)\right] - \varepsilon^2\frac{\sigma_M}{4}  - \sqrt{\frac{2\kappa N_{\rho\varepsilon} K}{N}\log\frac{2}{\delta}}. \\
\end{aligned}    
\end{equation}

Finally, we show that such distribution mismatch induces label noise in the adversarially augmented training set. Similar to the proof for the true classifier, by the common labeling practice of adversarial examples we have $P(\tilde{Y}'|x') = P(\tilde{Y}|x) = P(Y|x)$. 
By utilizing Lemma~\ref{theorem:implicit-label-noise}~\footnote{Note this is a result only associated with the training set, thus is not dependent on the specific classifier. } we then have with probability $1-3\delta$, 

\begin{equation}
    p_e(\mathcal{D}') \ge
    \varepsilon \left[(1 - \mathbb{E}_x f_\theta(x)_y) \frac{\sigma_m}{2L_\theta}- 2\rho\left(\left(\frac{3}{2} - \frac{1}{K}\right) L_\theta + L\right)\right]  - \varepsilon^2\frac{\sigma_M}{4}  
    - \xi \sqrt{\frac{1}{2N} \log\frac{2}{\delta}},
\end{equation}
where $\xi = 1 + \sqrt{4\kappa N_{\rho\varepsilon} K}$.
    

\end{proof}

\subsection{Proofs in Section~\ref{sect:mitigate-double-descent}}

\smallsection{Proof of Theorem~\ref{theorem: model-probability}}
\begin{proof}

Let $j^* = \text{argmax}~P(Y'=j|x')$ and thus $P(Y'=j^*|x') \in [1/c,1]$. Let $g(T) := f(x'; \theta, T)_{j^*}$, which is a continuous function defined on $[0, \infty]$. The condition $j^* = \argmax_j f(x'; \theta, T)_j$ ensures that $g(T) \in [1/c, 1]$, where $c$ is the number of classes. By the intermediate value theorem, there exists $T^*$, such that $g(T^*) = P(Y'=j^*|x')$.

Let $T = T^*$, we have
$$
\begin{aligned}
\|f(x'; \theta, T) - P(Y' | x')\|_{TV} 
& = \frac{1}{2} \sum_j \left|f(x'; \theta, T)_j - P(Y' =j | x')\right|\\
& = \frac{1}{2} \sum_{j, j\ne j^*} \left|f(x'; \theta, T)_j - P(Y' = j| x')\right|\\
& \le \frac{1}{2} \left[ \sum_{j, j\ne j^*} f(x'; \theta, T)_j + \sum_{j, j\ne j^*} P(Y' = j| x') \right]\\
& = 1 - P(Y'=j^*|x'),\\
\end{aligned}
$$
where the inequality holds by the triangle inequality.


Meanwhile, we have
$$
\begin{aligned}
\|P(\tilde{Y}' | x') - P(Y' | x')\|_{TV} 
& = \|P(Y | x) - P(Y' | x')\|_{TV}\\ 
& = \|\mathbbm{1}(y) - P(Y' | x')\|_{TV} \\
& = \frac{1}{2} \left[1 - P(Y' = y| x') + \sum_{j,j\ne \hat{y}} P(Y' = y| x')\right]\\
& = 1 - P(Y' = y| x')\\
& \ge 1 - P(Y' = j^* | x').\\
\end{aligned}
$$
Therefore, it can seen that for $T=T^*$,
$$
\| f(x'; \theta, T) - P(Y' | x') \|_{TV} \le \|  P(\tilde{Y}' | x') - P(Y' | x')\|_{TV}.
$$
\end{proof}

\smallsection{Proof of Theorem~\ref{theorem: model-probability-coefficient}}

\begin{lemma}
\label{lemma: model-probability-coefficient}
Let $x'$ be an example incorrectly classified by a classifier $f$ in terms of the true label distribution $P(Y'=j|x')$, namely
$$
\argmax_j~f(x'; \theta, T)_j \ne j^*,
$$
where $j^* = \argmax_j ~P(Y'=j|x')$. Assume $P(Y' = j^* | x') \ge 1/2$,
then 
$$ f(x'; \theta, T)_{j^*}  \le P(Y' = j^* | x').$$
\end{lemma}

\begin{proof}
We prove it by contradiction.
Assume $f(x'; \theta, T)_{j^*}  > P(Y' = j^* | x')$, we have $f(x'; \theta, T)_{j^*} > P(Y' = j^* | x') \ge 1/2$. Therefore,
$$
f(x'; \theta, T)_j \le \sum_{j, j\ne j^*} f(x'; \theta, T)_j = 1 - f(x'; \theta, T)_{j^*} < 1/2, ~ \forall j \ne j^*,
$$
which means $f(x'; \theta, T)_j < f(x'; \theta, T)_{j^*}, ~\forall j\ne j^*$. This leads to $j^* = \argmax_j f(x'; \theta, T)_j$, which contradicts our condition.
\end{proof}

Now we prove Theorem~\ref{theorem: model-probability-coefficient}
\begin{proof}

First let $P(Y'|x') = P(y|x) \approx \mathbbm{1}(y)$. 
Let $j^* = \argmax_j P(Y'=j|x')$. By Lemma \ref{lemma: model-probability-coefficient} we have $f(x'; \theta, T)_{j^*} \le P(Y'^*=j^*|x') \le 1$. Then there exists $\lambda^* > 0$, such that $\lambda^* \cdot f(x'; \theta, T)_{j^*} + (1 - \lambda^*) = P(Y'=j^* | x')$ by the intermediate value theorem.

Let $\lambda = \lambda^*$, we have
$$
\begin{aligned}
& 2 \left[ \|\lambda \cdot f(x'; \theta, T) + (1 - \lambda) \cdot P(\tilde{Y}' | x')- P(Y' | x')\|_{TV} - \| f(x'; \theta, T) - P(Y' | x')\|_{TV} \right]\\
= & 2 \left[ \|\lambda \cdot f(x'; \theta, T) + (1 - \lambda) \cdot \mathbbm{1}(y)- P(Y' | x')\|_{TV} - \| f(x'; \theta, T) - P(Y' | x')\|_{TV} \right]\\
= & \sum_j |\lambda \cdot f(x'; \theta, T)_j + (1 - \lambda) \cdot 1(j=y) - P(Y' =j | x')| - \sum_j |f(x'; \theta, T)_j - P(Y' =j | x')|\\
= & \sum_j |\lambda \cdot f(x'; \theta, T)_j + (1 - \lambda) \cdot 1(j=Y) - P(Y' =j | x')| - \sum_j |f(x'; \theta, T)_j - P(Y' =j | x')|\\
= & \sum_{j, j\ne j^*} |\lambda \cdot f(x'; \theta, T)_j - P(Y' = j | x')| - \sum_{j, j\ne j^*} |f(x'; \theta, T)_j - P(Y' = j | x')| - |f(x'; \theta, T)_{j^*} - P(Y' = j^* | x')|\\
\le & \sum_{j, j\ne j^*} |\lambda \cdot f(x'; \theta, T)_j - f(x'; \theta, T)_j| - |f(x'; \theta, T)_{j^*} - P(Y' = j^* | x')| \\
= & \sum_{j, j\ne j^*} [f(x'; \theta, T)_j - \lambda \cdot f(x'; \theta, T)_j ] - [P(Y' = j^* | x') - f(x'; \theta, T)_{j^*}] \\
= & \sum_{j, j\ne j^*} [f(x'; \theta, T)_j - \lambda \cdot f(x'; \theta, T)_j ] - [\lambda \cdot f(x'; \theta, T)_{j^*} + (1 - \lambda) - f(x'; \theta, T)_{j^*}] \\
= & \sum_j f(x'; \theta, T)_j - \lambda \sum_j f(x'; \theta, T)_j - (1-\lambda)\\
= & ~ 0.
\end{aligned}
$$
\end{proof}

\section{Limitations}
\label{sect:limitation}
We note that alternative labeling of adversarial examples proposed in this paper is based on the fact that the predictive distribution of a classifier trained with empirical risk minimization can approximate the true label distribution of training examples. However, such approximation may not be accurate especially if the classifier is not carefully regularized during training. Post-training confidence calibration techniques such as temperature scaling and interpolation can only improve the approximation in terms of the entire training set, but cannot improve it in a sample-wise manner. How to learn the true label distribution of adversarial training examples during adversarial training more accurately remains an open problem.

Also, such alternative labeling also requires to train another independent classifier beforehand, which induces additional training cost. 
\section{More empirical analyses}

\subsection{Epoch-wise double descent is ubiquitous in adversarial training}
\label{sect:double-descent-reconcile}


In this section, we conduct extensive experiments with different model architectures, and learning rate schedulers to verify the connection between robust overfitting and epoch-wise double descent. The default experiment settings are listed in Appendix~\ref{sect: exp-double-descent} in detail.







\smallsection{Model capacity}
We modulate the capacity of the deep model by varying the widening factor of the Wide ResNet. To extend the lower limit of the capacity, we allow the widening factor to be less than $1$. In such case, the number of channels in each residual block is scaled similarly but rounded, and the number of channels in the first convolutional layer will be reduced accordingly to ensure the width monotonically increasing through the forward propagation. 

\begin{figure*}[!ht]
\centering
\begin{subfigure}[t]{.48\textwidth}
  \centering
  \includegraphics[width=.95\textwidth]{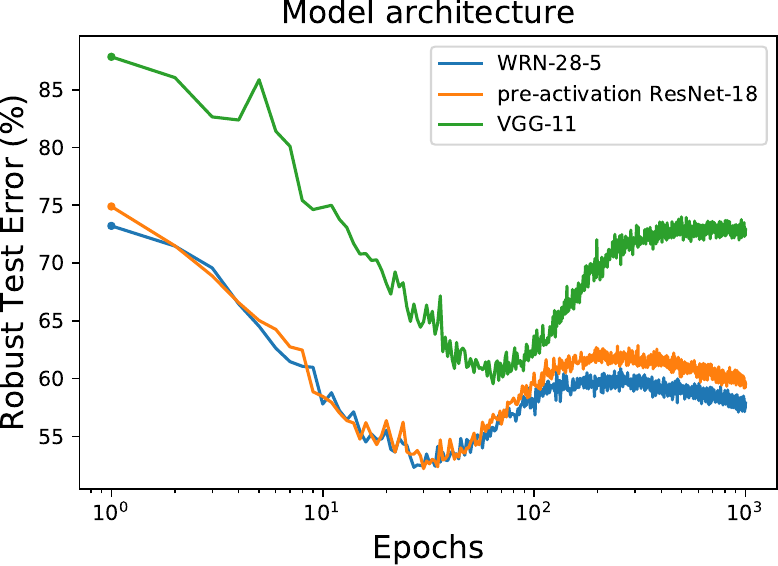}
  \caption{
  Epoch-wise double descent curves in adversarial training with various model architectures.
  }
  \label{fig:reconcile-model-architecture}
\end{subfigure}\hfill
\begin{subfigure}[t]{.48\textwidth}
  \centering
  \includegraphics[width=0.95\textwidth]{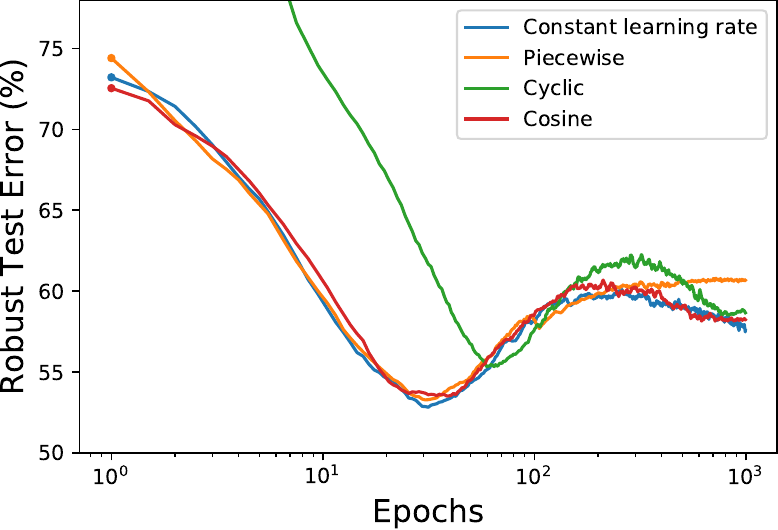}
  \caption{
   Epoch-wise double descent curves in adversarial training with various learning rate schedulers.
   The curves are smoothed by a moving average with a window of $5$ to avoid overlapping.
  }
\label{fig:reconcile-lr-scheduler}
\end{subfigure}
  \caption{Effect of model on the epoch-wise double descent curve}
 \label{fig:reconcile-model}
\end{figure*}

\smallsection{Model architecture} We also experiment on model architectures other than Wide ResNet, including pre-activation ResNet-18~\citep{He2016IdentityMI} and VGG-11~\citep{Simonyan2015VeryDC}. We select these configurations to ensure comparable model capacities\footnote{WRN-28-5, pre-activation ResNet-18 and VGG-11 have $9.13\times 10^6$, $11.17\times 10^6$ and $9.23\times 10^6$ parameters, respectively.}. As shown in Figure \ref{fig:reconcile-model}, different model architectures may produce slightly different double descent curves. The second descent of VGG-11 in particular will be delayed due to its inferior performance compared to residual architectures.



\smallsection{Learning rate scheduler}
A specific learning rate scheduler may shape the robust overfitting differently as suggested by \citet{Rice2020OverfittingIA}. We consider the following learning rate schedulers in our experiments.
\begin{itemize}[leftmargin=*]
    \item \textbf{Piecewise decay}: The initial learning rate rate is set as $0.1$ and is decayed by a factor of $10$ at the $100$th and $500$th epochs within a total of $1000$ epochs.
    \item \textbf{Cyclic}: This scheduler was initially proposed by \citet{Smith2017CyclicalLR} and has been popular in adversarial training. We set the maximum learning rate to be $0.2$, and the learning rate will linearly increase from $0$ to $0.2$ for the initial $400$ epochs and decrease to $0$ for the later $600$ epochs.
    \item \textbf{Cosine}: This scheduler was initially proposed by \citet{Loshchilov2017SGDRSG}. The learning rate starts at $0.1$ and gradually decrease to $0$ following a cosine function for a total of $1000$ epochs.
\end{itemize}
Experiments on various learning rate schedulers show the second descent can be widely observed except the piecewise decay, where the appearance of second descent might be delayed due to extremely small learning rate in the late stage of training. 






\section{More experiment results}


\subsection{Training longer}
\begin{figure*}[t]
  \centering
  \includegraphics[width=0.98\linewidth]{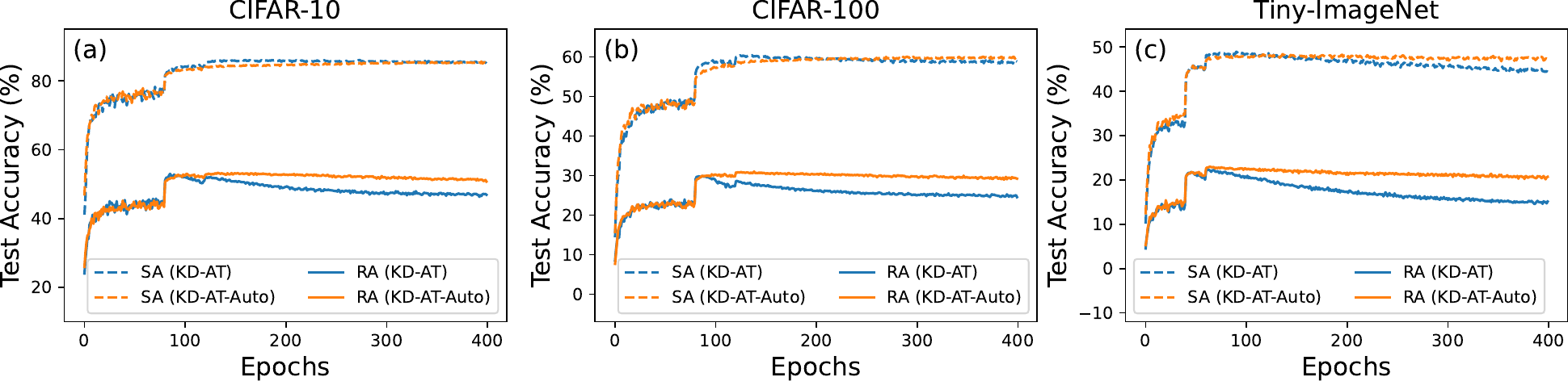}
  \vspace{-1ex}
  \caption{Our method can maintain robust test accuracy for more training epochs. 
  }
\label{fig:Longer-training}
\end{figure*}

As shown in Figure~\ref{fig:Longer-training}, We show that our method can maintain the robust test accuracy with more training epochs. Here, we follow the settings in Figure~\ref{fig:mitigate-overfitting} except we train for additional epochs up to $400$ epochs for each dataset.

\subsection{Adversarial training methods, neural architectures and evaluation metrics}
\label{sect:more-result}

In this section we conduct extensive experiments with different neural architectures, adversarial training methods and robustness evaluation metrics to verify the effectiveness of our method.

\begin{table*}[!ht]
  \small
  \caption{Performance of our method with different neural architectures.
   }
  \vspace{0.5ex}
  \label{table:result-model}
  \centering
  \small
  \begin{tabular}{clcccccccc}
    \toprule
    \multirow{2}{*}{Architecture} & \multirow{2}{*}{Setting} & \multirow{2}{*}{$T$} & \multirow{2}{*}{$\lambda$} & \multicolumn{3}{c}{Robust Acc. (\%)} & \multicolumn{3}{c}{Standard Acc. (\%)}\\
     & & &  & Best & Last & Diff. & Best & Last & Diff.\\
    \midrule
\multirow{3}{*}{VGG-19} 
& AT & - & - &  $42.21$ & $39.12$ & $ 3.09$ &  $73.95$ &  $80.45$ & -$6.50$ \\ 
& KD-AT & $2$ & $0.5$ &  $43.59$ & $42.69$ & $ 0.90$ &  $74.30$ &  $\textbf{77.80}$ & -$3.50$ \\ 
& KD-AT-Auto & $1.28^*$ & $0.79^*$ &  $\textbf{44.27}$ & $\textbf{44.24}$ & $ \textbf{0.03}$ &  $\textbf{76.41}$ &  $76.79$ & $\textbf{-0.38}$ \\ 
    \midrule
\multirow{3}{*}{WRN-28-5} 
& AT & - & - &  $49.85$ & $42.89$ & $ 6.96$ &  $84.82$ &  $85.87$ & -$1.05$ \\ 
& KD-AT & $2$ & $0.5$ &  $51.08$ & $48.40$ & $ 2.68$ &  $85.36$ &  $\textbf{86.88}$ & -$1.52$ \\ 
& KD-AT-Auto & $1.6^*$ & $0.82^*$ &  $\textbf{51.47}$ & $\textbf{51.10}$ & $\textbf{0.37}$ &  $\textbf{86.05}$ &  $86.24$ & $\textbf{-0.19}$ \\ 
    \midrule
\multirow{3}{*}{WRN-34-10} 
& AT & - & - &  $52.29$ & $46.04$ & $ 6.25$ &  $86.57$ &  $86.75$ & -$0.18$ \\ 
& KD-AT & $2$ & $0.5$ &  $53.11$ & $50.97$ & $ 2.14$ &  $86.41$ &  $\textbf{88.06}$ & -$1.65$ \\ 
& KD-AT-Auto & $1.6^*$ & $0.83^*$ &  $\textbf{54.17}$ & $\textbf{53.71}$ & $\textbf{0.46}$ &  $\textbf{87.69}$ &  $88.01$ & $\textbf{-0.32}$ \\ 
    \bottomrule
  \end{tabular}
\end{table*}

\begin{table*}[!ht]
  \small
  \caption{Performance of our method with different adversarial training methods. 
   }
  \vspace{0.5ex}
  \label{table:result-method}
  \centering
  \small
  \begin{tabular}{clcccccccc}
    \toprule
    \multirow{2}{*}{Method} & \multirow{2}{*}{Setting} & \multirow{2}{*}{$T$} & \multirow{2}{*}{$\lambda$} & \multicolumn{3}{c}{Robust Acc. (\%)} & \multicolumn{3}{c}{Standard Acc. (\%)}\\
     & & &  & Best & Last & Diff. & Best & Last & Diff.\\
    \midrule
\multirow{3}{*}{TRADES} 
& AT & - & - &  $48.50$ & $45.53$ & $ 2.97$ &  $82.79$ &  $82.68$ & $ 0.11$ \\ 
& KD-AT & $2$ & $0.5$ &  $48.74$ & $47.52$ & $ 1.22$ &  $82.30$ &  $\textbf{83.03}$ & -$0.73$ \\ 
& KD-AT-Auto & $1.12^*$ & $0.82^*$ &  $\textbf{48.75}$ & $\textbf{48.39}$ & $ \textbf{0.36}$ &  $\textbf{82.44}$ &  $82.80$ & $\textbf{-0.36}$ \\ 
    \midrule
\multirow{3}{*}{FGSM}
& AT & - & - &  $41.96$ & $35.39$ & $ 6.57$ &  $85.91$ &  $87.20$ & -$1.29$ \\ 
& KD-AT & $2$ & $0.5$ &  $42.82$ & $41.61$ & $ 1.21$ &  $86.69$ &  $\textbf{87.93}$ & -$1.24$ \\ 
& KD-AT-Auto & $2.18^*$ & $0.78^*$ &  $\textbf{44.11}$ & $\textbf{43.75}$ & $ \textbf{0.36}$ &  $\textbf{87.38}$ &  $87.66$ & $\textbf{-0.28}$ \\ 

    \bottomrule
  \end{tabular}
\end{table*}

\begin{table*}[!ht]
  \small
  \caption{Performance of our method under different adversarial attacks. PGD-1000 refers to PGD attack with $1000$ attack iterations, with step size fixed as $2/255$ as recommended by~\citet{Croce2020ReliableEO}.
   }
  \vspace{0.5ex}
  \label{table:result-evaluation}
  \centering
  \small
  \begin{tabular}{clccccc}
    \toprule
    \multirow{2}{*}{Attacks} & \multirow{2}{*}{Setting} & \multirow{2}{*}{$T$} & \multirow{2}{*}{$\lambda$} & \multicolumn{3}{c}{Robust Acc. (\%)} \\
     & & &  & Best & Last & Diff. \\
\multirow{3}{*}{PGD-1000} 
& AT & - & - &  $50.64$ & $43.00$ & $ 7.64$ \\
 & KD-AT & $2$ & $0.5$ &  $51.79$ & $48.43$ & $ 3.36$ \\ 
 & KD-AT-Auto & $1.47^*$ & $0.8^*$ &  $\textbf{52.05}$ & $\textbf{51.71}$ & $\textbf{0.34}$ \\ 
    \midrule
\multirow{3}{*}{Square Attack}
& AT & - & - & $53.47$ & $48.90$ & $4.57$ \\ 
& KD-AT & $2$ & $0.5$ &  $54.39$ & $52.92$ & $ 1.47$ \\ 
& KD-AT-Auto & $1.28^*$ & $0.79^*$ & $\textbf{55.23}$ & $\textbf{55.17}$ & $\textbf{0.06}$\\ 
    \midrule
\multirow{3}{*}{RayS}
& AT & - & - &  $55.76$ & $51.63$ & $ 4.13$ \\ 
& KD-AT & $2$ & $0.5$ & $56.59$ & $55.50$ & $ 1.09$ \\ 
& KD-AT-Auto & $1.6^*$ & $0.82^*$ &  $\textbf{57.74}$ & $\textbf{57.54}$ & $\textbf{0.20}$ \\ 

    \bottomrule
  \end{tabular}
\end{table*}

\FloatBarrier

\subsection{Combined with additional orthogonal techniques}
\label{sect:additional-technique}



We note that motivated from our theoretical analyses, our proposed method (KD-AT-Auto) is essentially the baseline knowledge distillation for adversarial training (KD-AT) with a robustly trained self-teacher, equipped with an algorithm that automatically finds its optimal hyperparameters (i.e. the temperature $T$ and the interpolation ratio $\lambda$). Stochastic Weight Averaging (SWA) and additional standard teachers (KD-Std) employed in~\citep{chen2021robust} are orthogonal contributions. KD-AT-Auto can certainly be combined with SWA and KD-Std to achieve better performance. 

As shown in Table~\ref{table:result-technique}, on CIFAR-10, KD-AT + KD-Std + SWA~\citep{chen2021robust} can already reduce the overfitting gap (difference between the best and last robust accuracy) to almost $0$. 
It is thus hard to see any further reduction by combining our method. 
To this end, we introduce an extra dataset SVHN~\citep{Netzer2011ReadingDI}. As shown in Table~\ref{table:result-technique}, on SVHN, KD-AT + KD-Std + SWA still produces a high overfitting gap (also see Appendix A1.3 in~\citep{chen2021robust}), whereas by combining with our algorithm to automatically find the optimal hyper-parameters (KD-AT-Auto + KD-Std + SWA), the overfitting gap can be further reduced to almost $0$. This demonstrates the effectiveness and wide applicability of our principle-guided method on mitigating robust overfitting. 


\begin{table*}[!ht]
  \small
  \caption{Performance of our method combined with SWA and an additional standard teacher. 
   }
  \vspace{0.5ex}
  \label{table:result-technique}
  \centering
  \small
  \scalebox{0.9}{
  \begin{tabular}{clcccccccc}
    \toprule
    \multirow{2}{*}{Dataset} & \multirow{2}{*}{Setting} & \multirow{2}{*}{$T$} & \multirow{2}{*}{$\lambda$} & \multicolumn{3}{c}{Robust Acc. (\%)} & \multicolumn{3}{c}{Standard Acc. (\%)}\\
     & & &  & Best & Last & Diff. & Best & Last & Diff.\\
    \midrule
    \multirow{3}{*}{CIFAR-10} 
    & AT & - & - &  $47.35$ & $41.42$ & $ 5.93$ &  $82.67$ &  $84.91$ & -$2.24$ \\
    &  KD-AT + KD-Std + SWA & $2$ & $0.5$ & $49.98$ & $49.89$ & $0.09$ & $\textbf{85.06}$ & $\textbf{85.52}$ & -$0.46$\\
    & KD-AT-Auto + KD-Std + SWA & $1.47^*$ & $0.8^*$ & $\textbf{50.03}$ & $\textbf{50.05}$ & $\textbf{-0.02}$ & $84.69$ & $84.91$ & $\textbf{-0.22}$\\ 
        \midrule
    \multirow{3}{*}{SVHN} 
    & AT & - & - & $47.83$ & $39.77$ & $8.06$ & $90.18$ & $91.11$ & -$0.93$\\
    & KD-AT + KD-Std + SWA & $2$ & $0.5$ & $47.88$ & $46.46$ & $1.42$ & $\textbf{91.59}$ & $\textbf{91.76}$ & $\textbf{-0.17}$\\
    & KD-AT-Auto + KD-Std + SWA  & $1.53^*$ & $0.83^*$ & $\textbf{50.58}$ & $\textbf{50.09}$ & $\textbf{0.49}$ & $90.54$ & $90.76$ & -$0.22$\\ 
    \bottomrule
  \end{tabular}
  }
\end{table*}

Here, the interpolation ratio of the standard teacher is fixed as $0.2$ and the SWA starts at the first learning rate decay for all experiments. We employ PGD-AT~\citep{Madry2018TowardsDL} as the base adversarial training method and conduct experiments with a pre-activation ResNet-18. The robust accuracy is evaluated with AutoAttack. Other experiment details are in line with Appendix~\ref{sect: exp-practical}.

Furthermore, we note that~\citep{chen2021robust} shows SWA and KD-Std are essential components to mitigate robust overfitting on top of KD-AT, while we show that KD-AT itself can mitigate robust overfitting by proper parameter tuning. We are thus able to separate these components and allow a more flexible selection of hyperparameters in diverse training scenarios without fear of overfitting. In particular, although~\citep{chen2021robust} suggests SWA starting at the first learning rate decay (exactly when the overfitting starts) mitigates robust overfitting, the effectiveness of SWA on mitigating overfitting may strongly depend on its hyper-parameter selection including $s_0$, i.e., the starting epoch and $\tau$, i.e., the decay rate\footnote{SWA can be implemented using an exponential moving average $\theta'$ of the model parameters $\theta$ with a decay rate $\tau$, namely $\theta' \leftarrow \tau \cdot \theta' + (1-\tau) \cdot \theta$ at each training step~\citep{Rebuffi2021FixingDA}.}, which is also mentioned in recent work~\citep{Rebuffi2021FixingDA}. We also did some additional experiments on CIFAR-10 following the SWA setting in~\citep{Rebuffi2021FixingDA} to demonstrate the wide applicability of our method. As shown by Table~\ref{table:result-swa}, when changing the hyperparameters of SWA, KD-AT + KD-Std + SWA cannot consistently mitigate robust overfitting, while KD-AT-Auto + KD-Std + SWA can maintain an overfitting gap close to 0 and achieve better robustness as well.

\begin{table*}[!ht]
  \small
  \caption{Performance of our method combined with SWA with different hyper-parameters}
  \vspace{0.5ex}
  \label{table:result-swa}
  \centering
  \small
  \begin{tabular}{lcccccccc}
    \toprule
    \multirow{2}{*}{Setting} & \multirow{2}{*}{$s_0$} & \multirow{2}{*}{$\tau$} & \multicolumn{3}{c}{Robust Acc. (\%)} & \multicolumn{3}{c}{Standard Acc. (\%)}\\
    & &  & Best & Last & Diff. & Best & Last & Diff.\\
    \midrule
    KD-AT + KD-Std + SWA & $80$ & $0.999$ & $49.00$ & $48.04$ & $0.96$ & $84.04$ & $\textbf{86.11}$ & -$2.07$\\
    KD-AT-Auto + KD-Std + SWA & $80$ & $0.999$ & $\textbf{49.35}$ & $\textbf{49.25}$ & $\textbf{0.1}$ & $\textbf{85.38}$ & $85.91$ & $\textbf{-0.37}$\\
    \midrule
    KD-AT + KD-Std + SWA & $0$ & $0.999$ & $49.01$ & $48.01$ & $1.0$ & $83.78$ & $\textbf{86.20}$ & -$2.42$\\
    KD-AT-Auto + KD-Std + SWA & $0$ & $0.999$ & $\textbf{49.32}$ & $\textbf{49.25}$ & $\textbf{0.07}$ & $\textbf{84.78}$ & $85.48$ & $\textbf{-0.7}$\\ 
    \bottomrule
  \end{tabular}
\end{table*}

\section{Study on a synthetic dataset with known true label distribution}



    
    \begin{wrapfigure}{r}{3cm} 
      \vspace{-3mm}
      \centering
      \includegraphics[width=0.2\textwidth]{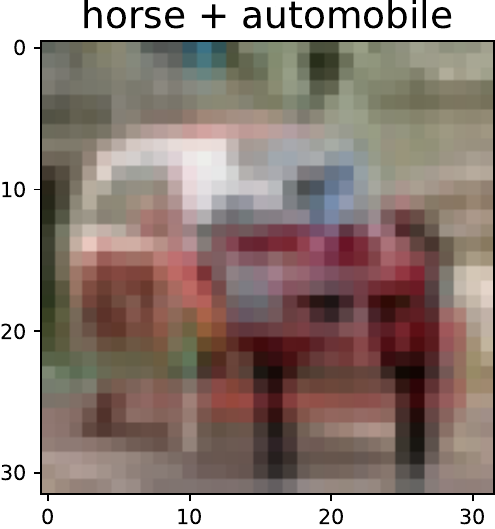}
      \caption{Sample image by mixup augmentation.}
    \label{fig:method-augment-example}
    \end{wrapfigure}
    
    \smallsection{Synthetic Dataset}
    Since the true label distribution is typically unknown for adversarial examples in real-world datasets, 
    we simulate the mechanism of implicit label noise in adversarial training from a feature learning perspective.
    Specifically, we adapt \emph{mixup}~\citep{Zhang2018mixupBE} for data augmentation on CIFAR-10. 
    For every example $x$ in the training set, we randomly select another example $x'$ in a different class and linearly interpolate them by a ratio $\rho$, namely $x:= \rho x + (1-\rho) x'$, which essentially perturbs $x$ with features from other classes. 
    Therefore, the true label distribution is arguably $y \sim \rho\cdot \mathbbm{1}(y) + (1-\rho)\cdot  \mathbbm{1}(y')$.
    Unlike mixup, we intentionally set the assigned label as $\hat{y} \sim \mathbbm{1}(y)$, thus deliberately create a mismatch between the true label distribution and the assigned label distribution.
    We refer this strategy as \emph{mixup augmentation} and only perform it once before the training. 
    In this way, the true label distribution of every example in the synthetic dataset is fixed.

\smallsection{Concentration of optimal temperature and interpolation ratio of individual examples}
\label{sect:optimal-temperature-mixup}
In Section~\ref{sect:approximate-true-distribution} we have shown that in terms of individual examples, the rectified model probability can provably reduce the distribution mismatch between the assigned label distribution and true label distribution of the adversarial example. 
However, since the true label distribution is unknown in realistic scenarios, it is not possible to directly follow Theorems~\ref{theorem: model-probability} and \ref{theorem: model-probability-coefficient} and calculate the optimal set of hyper-parameters for each example in the training set. The best we can do is to employ a validation set and determine a universal set of hyper-parameters based on the NLL loss, which expects all training examples to share similar optimal temperatures and interpolation ratios. 
Here, based on the synthetic dataset where a true label distribution is known, we empirically verify this assumption is reasonable.

\begin{figure*}[!ht]
  \centering
  \includegraphics[width=0.95\textwidth]{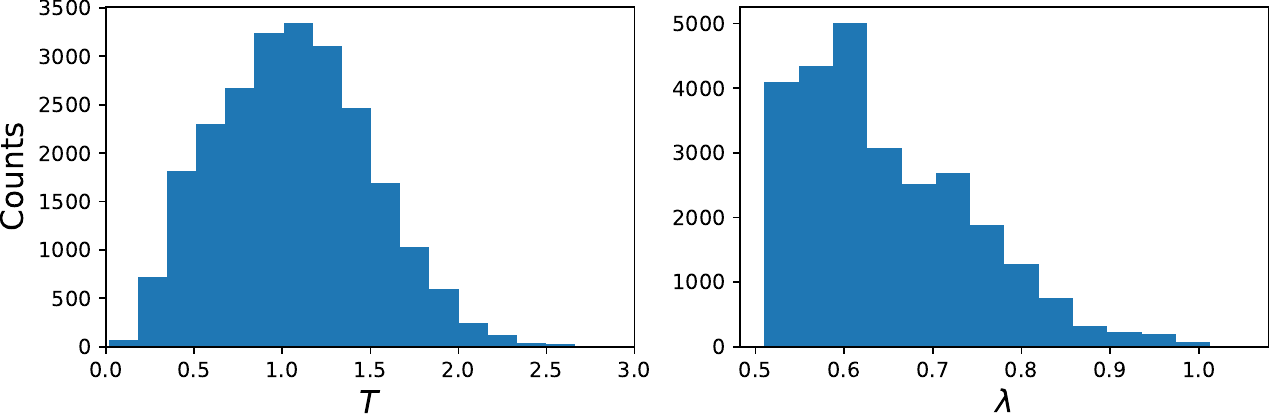}
  \caption{The histograms of optimal temperature (left) and interpolation ratio (right) of individual examples.
  }
\label{fig:method-augment-optimal}
\end{figure*}

In Figure~\ref{fig:method-augment-optimal} left, we solve the optimal temperature for each correctly classified training example based on Theorem~\ref{theorem: model-probability} with the interpolation ratio fixed as $1.0$. One can find that the individual optimal temperatures mostly concentrate between $0.5$ and $1.5$. In Figure~\ref{fig:method-augment-optimal} right, we solve the optimal interpolation ratio for each incorrectly classified training example based on Theorem~\ref{theorem: model-probability-coefficient} with the temperature fixed as $1.0$ . One can find that the individual optimal interpolation ratio mostly concentrate between $0.5$ and $0.7$.



\section{Method details}

\subsection{Determine the optimal hyper-parameters}
\label{sect:confidence-calibration-assigned}

One may note that Equation~(\ref{eq:calibration}) cannot be directly optimized since the traditional adversarial label is only defined on the example in the training set and cannot be simply generalized to the validation set.
A reasonable solution is using the nearest neighbour classifier to find the closest traditional adversarial label for every example in the validation set. 
However, to speed up the optimization, we propose to employ the classifier overfitted by the traditional adversarial labels on the training set as an surrogate, which works well in practice.
Specially, we employ a model overfitted on the training set to generate approximate traditional adversarial label of the adversarial example in the validation set. Such overfitted model is typically the model at the final checkpoint when conducting regular adversarial training for sufficient epochs. Mathematically, our final method to determine the optimal temperature and interpolation ratio in rectified model probability can be described as
\begin{equation}
    T, \lambda = \argmin_{T, \lambda} \mathbbm{E}_{(x', y')\sim \mathcal{D}'_\text{val}}~\ell\left(\lambda\cdot f_\theta(x'; T) + (1-\lambda)\cdot f_{\theta_s}(x'; T), y'\right),
\end{equation}
where $f_{\theta_s}(x'; T)$ denotes the temperature-scaled predictive probability of a surrogate model on $x'$. Here the validation set is constructed by applying adversarial perturbation generated by $f_\theta$ to the clean validation set. For adversarial perturbation we utilize PGD attack with $10$ iterations, the perturbation radius as $8/255$ and the step size as $2/255$.
Note that Such process incurs almost no additional computation as we simply obtain the logits of a surrogate classifier. 

\section{Experimental details}
\label{sect:exp-setting-all}

\subsection{Settings for main experiment results}
\label{sect: exp-practical}
\smallsection{Dataset}
We include experiment results on CIFAR-10, CIFAR-100, Tiny-ImageNet and SVHN.

\smallsection{Training setting}
We employ SGD as the optimizer. The batch size is fixed to 128. The momentum and weight decay are set to $0.9$ and $0.0005$ respectively. Other settings are listed as follows.
\begin{itemize}[leftmargin=*,nosep]
    \item CIFAR-10/CIFAR-100: we conduct the adversarial training for $160$ epochs, with the learning rate starting at $0.1$ and reduced by a factor of $10$ at the $80$ and $120$ epochs.
    \item Tiny-ImageNet: we conduct the adversarial training for $80$ epochs, with the learning rate starting at $0.1$ and reduced by a factor of $10$ at the $40$ and $60$ epochs.
    \item SVHN: we conduct the adversarial training for $80$ epochs, with the learning rate starting at $0.01$ (as suggested by \citep{chen2021robust}) and reduced by a factor of $10$ at the $40$ and $60$ epochs.
\end{itemize}

\smallsection{Adversary setting}
We conduct adversarial training with $\ell_\infty$ norm-bounded perturbations. We employ adversarial training methods including PGD-AT, TRADES and FGSM. We set the perturbation radius to be $8/255$. For PGD-AT and TRADES, the step size is $2/255$ and the number of attack iterations is $10$.

\smallsection{Robustness evaluation}
We consider the robustness against $\ell_\infty$ norm-bounded adversarial attack with perturbation radius $8/255$. We employ AutoAttack for reliable evaluation. We also include the evaluation results again PGD-1000, Square Attack and RayS.

\smallsection{Neural architectures}
We include experiments results on pre-activation ResNet-18, WRN-28-5, WRN-34-10 and VGG-19. 

\smallsection{Hardware}
We conduct experiments on 
NVIDIA Quadro RTX A6000.

\subsection{Settings for analyzing double descent in adversarial training}
\label{sect: exp-double-descent}

\smallsection{Dataset}
We conduct experiments on the CIFAR-10 dataset, without additional data.

\smallsection{Training setting}
We conduct the adversarial training for $1000$ epochs unless otherwise noted.
By default we use SGD as the optimizer with a fixed learning rate $0.1$. When we experiment on a subset (see below) we use the Adam optimizer to improve training stability, where the learning rate is fixed as $0.0001$.
The batch size will be fixed to $128$, and the momentum will be set as $0.9$ wherever necessary. No regularization such as weight decay is used. These settings are mostly aligned with the empirical analyse of double descent under standard training~\citep{Nakkiran2020DeepDD}.

\smallsection{Sample size}
To reduce the computation load demanded by an exponential number of training epochs, we reduce the size of the training set by randomly sampled a subset of size $5000$ from the original training set without replacement. 
We adopt this setting for extensive experiments for analyzing the dependence of epoch-wise double descent on the perturbation radius and data quality (i.e. Figure~\ref{fig:dependence-perturbation-quality})..

\smallsection{Adversary setting}
We conduct adversarial training with $\ell_\infty$ norm-bounded perturbations. We employ standard PGD training with the perturbation radius set to $8/255$ unless otherwise noted. The number of attack iterations is fixed as $10$, and the perturbation step size is fixed as $2/255$.
 
\smallsection{Robustness evaluation}
We consider the robustness against $\ell_\infty$ norm-bounded adversarial attack with perturbation radius $8/255$. We use PGD attack with $10$ attack iterations and step size set to $2/255$.

\smallsection{Neural architecture}
By default we experiment on Wide ResNet~\citep{Zagoruyko2016WideRN} with depth $28$ and widening factor $5$ (WRN-28-5) to speed up training. 

\smallsection{Hardware}
We conduct experiments on 
NVIDIA Quadro RTX A6000.

\subsection{Estimation of the data quality}
In this section we elaborate on the calculation of data quality for analyzing the dependence on label noise in adversarial training.

We use the predicative probabilities of classifiers trained on CIFAR-10 to score its training data. Similar strategy is employed in previous works to select high-quality unlabeled data to improve adversarial robustness~\citep{Uesato2019AreLR, Carmon2019UnlabeledDI, Gowal2020UncoveringTL}. Slightly deviating from these works focusing on out-of-distribution data, we use adversarially trained instead of regularly trained models to measure the quality of in-distribution data, since under standard training almost all training examples will be overfitted and gain overwhelmingly high confidence. Specifically, we adversarially train a pre-activation ResNet-18 with PGD and select the model at the best checkpoint in terms of the robustness. The quality of an example is estimated by the model probability corresponding to the true label without adversarial perturbation and random data augmentation (flipping and clipping). We repeat this process $10$ times with random initialization to obtain a relatively accurate estimation.




\subsection{Settings for standard training on fixed augmented training sets}

\subsubsection{General settings for both adversarial augmentation and Gaussian augmentation}

\smallsection{Dataset}
We conduct experiments on the CIFAR-10 dataset, without additional data.

\smallsection{Training setting}
We conduct the standard training for $1000$ epochs.
We use Adam as the optimizer with a fixed learning rate $0.0001$ to improve training stability with a small training set (see below). The batch size will be fixed to $128$, and the momentum will be set as $0.9$ wherever necessary. No regularization such as weight decay is used.

\smallsection{Sample size}
To reduce the computation load demanded by an exponential number of training epochs, we reduce the size of the training set by randomly sampled a subset of size $5000$ from the original training set without replacement.

\smallsection{Neural architecture}
By default we experiment on Wide ResNet~\citep{Zagoruyko2016WideRN} with depth $28$ and widening factor $5$ (WRN-28-5).

\smallsection{Hardware}
We conduct experiments on 
NVIDIA Quadro RTX A6000.

\subsubsection{Construction of the training set}


\smallsection{Adversarial augmentation}
We first obtain a robust model by conduct PGD training with pre-activation ResNet-18 on CIFAR-10. We use early stopping to obtain the most robust model on a validation set. The specific settings are aligned with Section~\ref{sect: exp-practical}.

Using this model, we then generate adversarial examples with PGD attack on the $5000$ examples randomly sampled from CIFAR-10 training set. The number of attack iterations is fixed as $10$ and the step size is fixed as $2/255$. The adversarial examples along with their original labels are then grouped into a training set for adversarial augmentation experiments.

\smallsection{Gaussian augmentation}
We apply Gaussian noise to the $5000$ examples randomly sampled from CIFAR-10 training set. The perturbed examples along with their original labels are then grouped into a training set for Gaussian augmentation experiments. 

\end{document}